\documentclass[final,1p,times]{elsarticle}

\usepackage{amsthm,amsmath,amssymb}
\usepackage{amsfonts}
\usepackage{mathrsfs}
\usepackage{apalike}
\biboptions{authoryear}

\usepackage{url} 
\usepackage{cite}
\usepackage{amsthm,amsmath}
\usepackage{mathrsfs}
\usepackage{array}
\usepackage{tikz}
\usepackage[skins]{tcolorbox} 
\tcbuselibrary{breakable} 
\usepackage{multirow}
\usepackage{hyperref}
\usepackage[ruled,linesnumbered]{algorithm2e}
\usepackage{booktabs}  
\usepackage{amssymb}

\usepackage{colortbl}
\usepackage[capitalize]{cleveref}
\Crefname{section}{Section}{Sections}
\crefname{section}{Sec.}{Secs.}
\Crefname{figure}{Figure}{Figures}
\crefname{figure}{Fig.}{Figs.}
\Crefname{table}{Table}{Tables}
\crefname{table}{Tab.}{Tabs.}
\Crefname{algorithm}{Algorithm}{Algorithms}
\crefname{algorithm}{Alg.}{Algs.}

\usepackage[caption=false, font=normalsize, labelfont=sf, textfont=sf]{subfig}

\usepackage{etoolbox}
\makeatletter
\patchcmd{\@makecaption}
  {\scshape}
  {}
  {}
  {}
\makeatletter
\patchcmd{\@makecaption}
  {\\}
  {.\ }
  {}
  {}
\makeatother

\usepackage{flushend} 
\usepackage{tabularx} 
\usepackage{ragged2e} 
\newcolumntype{L}{>{\RaggedRight\hangafter=1\hangindent=0em}X}

\usepackage{xspace}
\makeatletter
\DeclareRobustCommand\onedot{\futurelet\@let@token\@onedot}
\def\@onedot{\ifx\@let@token.\else.\null\fi\xspace}

\def\eg{\emph{e.g}\onedot} 

\def\ie{\emph{i.e}\onedot}

\def\etc{\emph{etc}\onedot} 
\def\vs{\emph{vs}\onedot}

\def\iid{i.i.d\onedot}

\makeatother

\newcommand{\distr}{\phi}
\newcommand{\diste}{\psi}
\newcommand{\losskl}{\mathcal{L}_{KL}}
\newcommand{\lossce}{\mathcal{L}_{CE}}
\newcommand{\losskd}{\mathcal{L}_{reg}}

\newcounter{ToDo}
\newcounter{smcomm}
\newcounter{Note}
\definecolor{skyblue}{rgb}{0.53, 0.81, 0.92}
\definecolor{airforceblue}{rgb}{0.36, 0.54, 0.66}
\definecolor{blueryb}{rgb}{0.01, 0.28, 1.0}
\definecolor{aoen}{rgb}{0.0, 0.5, 0.0}
\definecolor{awesome}{rgb}{1.0, 0.13, 0.32}
\definecolor{americanrose}{rgb}{1.0, 0.01, 0.24}

\journal{Elsevier}

\begin{document}

\begin{frontmatter}
\title{\textsc{ShiftKD}: Benchmarking Knowledge Distillation under Distribution Shift}

\author[label1, label4]{Songming Zhang}
\ead{zhangsongming@suat-sz.edu.cn}

\author[label3]{Yuxiao Luo}
\ead{yuxiao.luo@connect.polyu.hk}

\author[label1]{Ziyu Lyu\corref{cor1}}
\ead{lvzy7@mail.sysu.edu.cn}

\author[label2]{Xiaofeng Chen}
\ead{xxffch@126.com}

\cortext[cor1]{Corresponding author}

\address[label1]{School of Cyber Science and Technology, Shenzhen Campus of Sun Yat-sen University}
\address[label2]{Department of Mathematics, Chongqing Jiaotong University, China}
\address[label3]{Department of Land Surveying and Geo-informatics, The Hong Kong Polytechnic University, Hong Kong}
\address[label4]{Shenzhen University of Advanced Technology}


\begin{abstract}
    Knowledge Distillation (KD) transfers knowledge from large models to small models and has recently achieved remarkable success. 
However, the reliability of existing KD methods in real-world applications, especially under distribution shift, remains underexplored. 
Distribution shift refers to the data distribution drifts between the training and testing phases, and this can adversely affect the efficacy of KD. 
In this paper, we propose a unified and systematic framework \textsc{ShiftKD} to benchmark KD against two general distributional shifts: diversity and correlation shift. 
The evaluation benchmark covers more than 30 methods from algorithmic, data-driven, and optimization perspectives for five benchmark datasets. 
Our development of \textsc{ShiftKD} conducts extensive experiments and reveals strengths and limitations of current SOTA KD methods.
More importantly, we thoroughly analyze key factors in student model training process, including data augmentation, pruning methods, optimizers, and evaluation metrics.
We believe \textsc{ShiftKD} could serve as an effective benchmark for assessing KD in real-world scenarios, thus driving the development of more robust KD methods in response to evolving demands.
The code will be made available upon publication.
\end{abstract}

\begin{highlights}
\item  ShiftKD: First Distribution Shift-Aware Knowledge Distillation Benchmark.
\item Systematic Evaluation of 30+ SOTA Methods.
\item Fine-grained Impact Analysis Across Shifted Domains and Conditions.
\end{highlights}

\begin{keyword}
Knowledge distillation \sep  distribution shift \sep  benchmark
\end{keyword}
\end{frontmatter}


\section{Introduction}
\label{sec:intro}
In recent decades, machine learning systems have typically focused on large-parameter visual networks and shown remarkable advancement. Despite their impressive performance, the rapid spread of end-side devices such as mobile devices, has been accompanied by two critical limitations. Firstly, the high computational requirements during the training of these networks hinder their widespread deployment in resource-constrained environments. Secondly, distribution shifts—situations where the data distribution during inference significantly differs from that during training~\citep{yang2023change}—can lead to performance degradation. When the model encounters a new data distribution, it may not generalize well, causing a discrepancy in performance. Consequently, many recent advancements in vision tasks have not been effectively translated into practical applications.

To address such a problem, we focus on the following tasks: Given a large, well-trained model, the goal is to compress it into a smaller, more robust structure without compromising its performance under distribution shift. Knowledge distillation (KD) is notable for its superior performance in model compression, which involves training smaller student models to mimic the behavior and knowledge of larger networks.
As KD is widely used in various fields~\citep{LE2025107017,TIAN2024106567}, a wealth of previous work is dedicated to understanding the basic mechanisms of KD and the generalization capabilities shared between teacher and student models, assuming independent and identically distribution (\iid case). \citet{stanton2021Does} suggest that the student may not perfectly match the teacher’s predictive distributions due to optimization challenges and the influence of distillation datasets, and closely mirroring the teacher does not always guarantee better generalization.
However, the efficacy of KD can be limited by distribution shift (non-\iid case), and the previous explanations are not sufficient to analyze the shift scenario. 

\begin{figure*}[t]
    \centering
    \includegraphics[width=\linewidth]{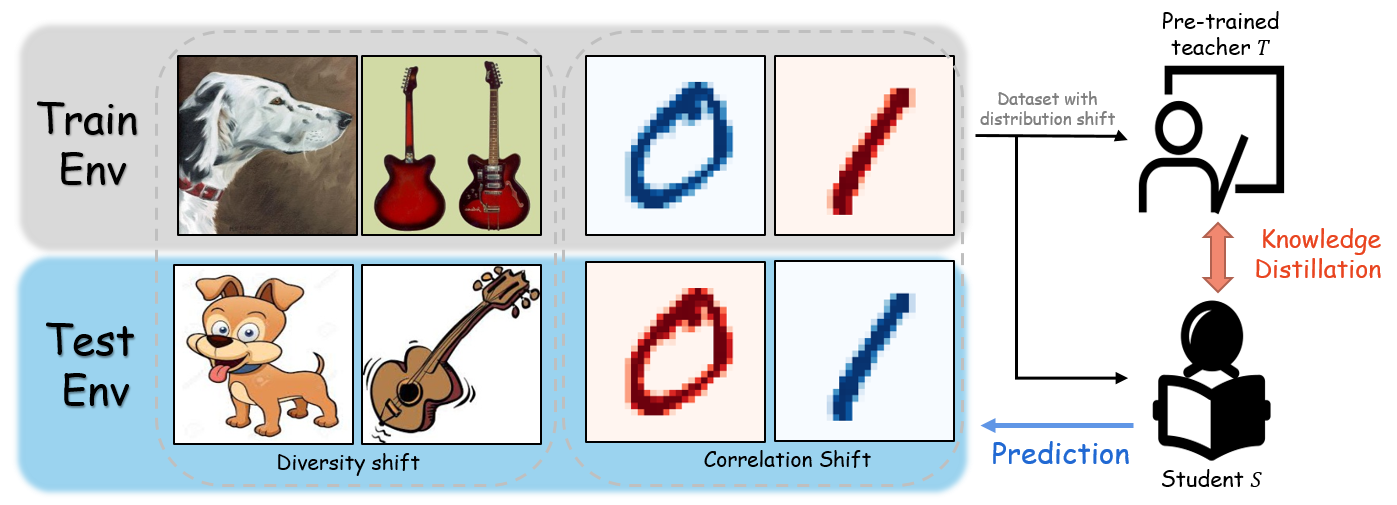}
    \caption{\textit{(Left)} Knowledge distillation tends to exhibit overfitting to distribution shift in real-world applications. Distribution changes are typically categorized as diversity and correlation shifts according to the changes in $P(X)$ and $P(Y | X)$, respectively, such as style-changed pictures or color-reversed digits. \textit{(Right)} We propose a unified framework to evaluate knowledge distillation under distribution shift and compare its effectiveness against more than 20 algorithms. The perspective of various algorithms is mainly from three levels, including algorithm, data, and optimization.}
    \label{fig:dog_ng}
\end{figure*}

Distribution shifts in real deployment may lead to a drop in the performance of the student model after distillation. For example, while a lightweight student learns to identify dogs or colored digits with the help of large models, it may not generalize well to real-world samples that it has not seen before.
As shown in \cref{fig:dog_ng}, the training environment often contains clean and well-aligned data, whereas the deployment environment may exhibit substantial distributional changes, such as diversity shifts (e.g., stylistic changes from photo-realistic to cartoon images) or correlation shifts (e.g., label-correlated cues like color being reversed at test time).
Such shifts create a trade-off between model capacity and generalization performance on the student mode~\citep{wang2021embracing}. However, little is known or has been evaluated about when and how well the student model learns from the teacher against distribution shift; most work has focused on improving KD performance. \citet{fang2021mosaicking} augment the in-domain distillation performance by incorporating out-of-domain data. \citet{zhou2022device} enhance out-of-domain generalization of KD through aggressive augmentation techniques, while \citet{do2022momentum} focus on data-free approaches to KD.  Unlike previous work, we investigate whether existing methods are effective solutions for distribution shifts. This naturally leads to the following research questions:
\begin{center}
    \begin{tcolorbox}[colback=gray!9, colframe=black, width=\columnwidth,
                  arc=0.8mm, auto outer arc, boxrule=0.4pt, 
                  title = Our empirical questions:]
\emph{ \textbf{Q1:} Do existing knowledge distillation methods remain effective under distribution shift? Specifically, how well do student models align with their teachers across various types of knowledge? \textbf{Q2:} Which data strategies, such as augmentation or pruning, are better for mitigating distribution shifts? \textbf{Q3:} What are the most effective optimization techniques in this context?}
    \end{tcolorbox}
\end{center}

To address these questions, we propose a KD paradigm tailored for distributional shift situations, \textsc{ShiftKD}. This novel paradigm involves rethinking traditional KD settings and reformulating the objectives to accommodate multiple domains in real-world scenarios. In line with this new paradigm, we identify and reconsider potentially influential factors from three distinct perspectives: algorithmic, data, and optimization levels. We also delineate two types of distribution shifts characterized by the properties of domain data, as previously described by~\citet{ye2022ood}. Furthermore, we introduce a comprehensive and systematic evaluation framework to benchmark the effectiveness of KD under distributional shift scenarios. The evaluation benchmark encompasses a broad spectrum of KD approaches categorizing over 30 methods. 
We conduct extensive experiments in our benchmark and show that this framework analyzes KD performance across various real-world settings, which is truly crucial. It helps understand how different strategies work and improves KD’s adaptability by addressing its weaknesses.

As an evaluative benchmark grounded in shifted constraints, \textsc{ShiftKD} demonstrates significant potential to support the development of more robust KD methods that better meet practical needs.
Moreover, our framework is easily extendable and introduces new dimensions to recently emerging areas. 

Specifically, the main contributions are as follows:
\begin{itemize}
    \item We formulate a novel knowledge distillation framework evaluated under distribution shift by rethinking previous \iid settings theoretically. Our framework considers influential factors and offers comprehensive insights for tackling such a problem during the distillation process.
    \item We establish a thorough and systematic evaluation benchmark for knowledge distillation in distribution shift situations, encompassing a wide spectrum of approaches with over 30 algorithms. We believe this to be the first work that evaluates the performance of various distillation methods against distribution shifts.
    \item We present constructive findings that explain when and how existing KD methods work against distribution shifts through average and worst-case scenarios. Few knowledge distillation methods are effective against all shift conditions, and the vanilla method is sometimes enough. It is noteworthy that the effectiveness of dark knowledge and data augmentation diminishes sharply when distribution shifts occur, a point never addressed in previous work. Data pruning and pre-training may be beneficial for the student model against distribution shifts, but this depends on the student's specific task.
\end{itemize}
\section{\textsc{ShiftKD}: Framework to Evaluate Knowledge Distillation to Distribution Shift}
\label{sec:framework}

\subsection{Preliminaries}
\label{sec:framework-pre}

\subsubsection{Knowledge Distillation (KD)} 
\citet{hinton2015distilling} first proposed KD to transfer knowledge from an elaborate complex network to a shallower one. 
Let $X$ and $Y$ be random variables taking values in $\mathcal{X}$ and $\mathcal{Y}$, respectively.
In the supervised classification setting, given a large-scale teacher model $T = T(X;\theta_t)$ optimized on training data $D_{tr}$.
The lightweight student model $S = S(X;\theta_s)$ directly minimizes the following objective:
\begin{equation} \label{eq:kd}
    \min_{\theta_s} \mathbb{E}[\alpha \losskl(S, T)+ (1-\alpha) \lossce(S, Y)]
\end{equation}
where the distillation weight $\alpha \in [0,1]$. $\losskl$ is the transfer loss term that encourages $S$ to imitate the predictive distribution of $T$, and $\lossce$ is the cross-entropy between the student output and the ground-truth labels.

\subsubsection{KD under distribution shift (non-i.i.d. case)} 
In the context of non-i.i.d. cases, we are given $K$ similar but distinct training domains $D_{tr}=\{D^e = (X^e, Y^e)\}_{e=1}^K$, each are sampled from the training environment as distribution $P^e_{XY} $. Note that $P^e_{XY}  \neq P^{e'}_{XY} , e \neq e'$ and $e, e' \in \{1, \dots, K\}$.
In general, \cref{eq:kd} works under the \iid assumption which does not hold in real scenarios, as \cref{fig:dog_ng}.
In the case of training distribution is not equal to test distribution (i.e., $P_{tr} \neq P_{te}$ with probability function $\distr \neq \diste$), the student might encounter failure, eventually not being applicable.
The goal of KD under distribution shift is to construct a student model $S(X;\theta_s)$ that can perform well in an unseen environment $D_{te}$. That is, $D_{te}$ is not accessible in training. 

Theoretically, we expect the student model to generalize to any test distribution $P^{te}$ that may have unseen shifts. 
However, in practice, the trade-off between handling transformations and model capacity is complex and involves many practical tricks, which limits the student model.
Motivated by this gap between theory and reality, the research question is: \textbf{There are many existing methods designed for knowledge distillation, do they still work under distribution shift?} 

To investigate this issue, this paper proposes a KD evaluation framework for distribution shifts.
In the framework, the teacher model $T$ is trained on a dataset $D_{tr}$ with distribution shifts and then distilled to the student $S$. The goal of the framework is to investigate and evaluate the performance on test set $D_{te}$ of student $S$ against changes in the distribution, that is, the robustness of the KD process. Even if the teacher $T$ is not robust, we still want to extract a robust student $S$ by KD.

\subsection{Framework Setting}
In the proposed evaluation framework, we investigate the effects of distribution shift on the distillation process, delving into distinct and complementary viewpoints in real cases, respectively KD algorithms \emph{(algorithm-level)}, data manipulation mechanisms \emph{(data-level)}, and optimization selection \emph{(optimization-level)}. Furthermore, we introduce the precise settings of three specific circumstances that arise under distribution shifts.

\subsubsection{Transferable Knowledge algorithms} 
Current KD algorithms can be classified based on the types of transferred knowledge.
While considering the different knowledge used in diverse algorithms, one can reformulate the goal as follows:
\begin{equation} \label{eq:kd_2}
    \min_{\theta_s} \mathbb{E}[\alpha \losskl(S, T)+ (1-\alpha) \lossce(S, Y) + \beta \losskd],
\end{equation}
where $\losskd$ denotes different knowledge sources adopted as the regularization terms of KD and $\beta$ is the trade-off hyperparameter.
Based on \cref{eq:kd_2}, we aim to explore the trends and reasons for the impact of different knowledge types under distribution shift. Specifically, we seek to answer the question: \textbf{What kinds of knowledge can help the student match teacher well against distribution shift?}
Thus, we analyze KD algorithms by categorizing them into three distinct categories, as proposed by~\citet{gou2021knowledge}. 
\begin{itemize}
    \item Logit-based knowledge, which is most popular and imitates the teacher's final layer directly (KD~\citep{hinton2015distilling}). This technique allows the student model to learn from the softened probabilities of the teacher model, which contain more information than the hard labels. It also reduces the overfitting problem of the student model by smoothing the output distribution.
    \item Feature-based knowledge, which refers to student works with specific intermediate layers in its teacher. The different concerns to the hint layers affect feature selection during transferring, such as attention maps. For example, AT~\citep{zagoruyko2016paying} enables students to mimic attention maps of a strong teacher network.
    \item Relation-based knowledge, which leverages the relevance of model layers or samples as information to guide student's learning, such as contrastive learning or similarity matrix. For example, SP~\citep{tung2019similarity} transfers the relative similarity information of the teacher model, and CRD~\citep{tian2019contrastive} transfers the instance-level contrastive information by using a discrimination objective.
    Knowledge of the relevance is informative and can be shared to guide learning. 
\end{itemize}

\subsubsection{Distillation Data Manipulation} 
The impact of distribution shift is not solely determined by distillation algorithms but is also significantly influenced by differences in data. 
In real-world scenarios, the teacher model trained on insufficient or inadequate data is often imperfect, and the performance of the student model is highly dependent on access to a large and high-quality training dataset. However, in the presence of distribution shift, acquiring such a dataset becomes an insurmountable challenge. 
It is natural to ask \textbf{how do we choose a proper data strategy to ensure the robustness of distillation to distribution shift?} 

For the situation of data manipulation-based KD, We reformulate the learning objective as:
\begin{equation} \label{eq:kd_1}
    \min_{\theta_s} \mathbb{E}[\alpha \ \losskl(S, T)+ (1-\alpha) \lossce(S^e, Y)],~~ (\hat{X},Y) \in P^{tr}_{XY},
\end{equation} By changing the input $x$ to $\hat{x}$ with manipulation function $M(\cdot)$, this technique can assist students in better learning.
In our framework, the quality and diversity of distilled data make us study two manipulation mechanisms, including data augmentation and pruning:

(1) If distillation data is the main cause of poor teaching performance, \textbf{data augmentation}, which is a simple and effective way to improve the coverage of data distribution and inter-domain robustness, \ie, $M(X)=\hat{X}$ and $ (\hat{X},Y) \in P^{tr}_{XY}$. We pay more attention to the augmentation approach based on randomization or generation. Random-based augmentation is typically achieved by creating new complex environments based on randomized manipulation. For example, AutoAugment~\citep{cubuk2018autoaugment} uses reinforcement learning to search for the optimal augmentation policies from a predefined search space for different datasets and tasks, and RandAugment~\citep{cubuk2020randaugment} simplifies the search process of AutoAugment. Additionally, generation-based augmentation concerns the creation of more diverse data at the feature level, \eg, Mixup~\citep{zhang2017mixup} a method that creates new images and labels by taking convex combinations of two images and their corresponding labels. For example, EL2N~\citep{paul2021deep} is a pruning method that assigns an importance score to each data point based on the output of model prediction and prunes the data points with low scores.

(2) What if there was no augmentation, and the student was distilled from important or representative samples? \textbf{Data pruning}~\citep{sorscher2022beyond}, which is the approach to quantifying individual sample differences by removing low-quality or immaterial samples in the dataset, \ie, $M(D_{tr})= \{ \hat{D}_{tr}=  (\hat{X},Y) \}$ and $ (\hat{X},Y) \in  P^{tr}_{XY}$. We expect to improve the training efficiency and robustness to distribution shifts by improving the quality of distillation data.

\subsubsection{Optimization option} 
Prior studies have shown that different optimization settings can lead to varying performance on student models in KD~\citep{stanton2021Does}. It is worth noting that our framework for distribution shift differs from previous studies due to the lack of access to the unknown test environment. KD involves several factors that inherently affect the training process, such as hyperparameter selection and the teacher-student architecture. To gain a better understanding of the influence of such factors, we conduct empirical observations with all other variables held constant. 

Specifically, we will focus on the following aspects:
(1) Hyperparameters are used to control the distillation process of the model. In the case of distribution shifts, the original hyperparameters may no longer be applicable to new data distribution, so we study the transfer effect of dark knowledge by focusing on hyperparameters. (2) Pretraining aims to allow the model to learn common feature representations. However, this can lead to poor performance of the pre-trained model on new distribution, especially for the teacher model. (3) The optimizer is a useful tool to update models. For the distillation process under distribution shift, the choice of optimizer may cause the student model to fall into a local optimum. (4) We evaluate the distillation process and the student model from different dimensions by using different performance metrics and try to further explore the transfer reliability of KD under distribution shift.

\subsubsection{Types of Distribution shift}  
To better study the performance of KD under distributional shift, we propose to characterize how features change in the downstream domain.
Given a training dataset $D_{tr}=\{D^e = (X^e, Y^e)\}_{e=1}^K$, each  are sampled from the joint distribution  $P^e_{XY}$. 
The input $X$ is determined by the semantic factor $Se$ and the variation factor $V$ based on causal theory, while the output $Y$ is determined by $Se$ independently. 
 Inspired by the real world, \citet{ye2022ood} formalize the distribution shift into two types, namely \emph{diversity shift} and \emph{correlation shift}.

\emph{Diversity shift} describes the fact that each environment in the dataset represents a diversity of characteristics in domains. For example, learning the images from the domain of art painting, but testing on the cartoon-style samples (such as \cref{fig:dog_ng} 
 Left).
\emph{Correlation shift} is caused by spurious correlations in the data that have received more recent attention. For example, the MNIST variant, ColorMNIST~\citep{arjovsky2019invariant}, consists of the digits with red or green, but flipping the strong correlation of colors and labels in different environments (such as \cref{fig:dog_ng} Right). 
The quantification formula of two shifts is defined to support the comprehension and evaluation of KD algorithms, respectively:
\begin{align}
        F_{div}(\distr, \diste) &= \frac{1}{2} \int_{\mathcal{V}_1} | \distr(v) - \diste(v) | dv \\
        F_{cor}(\distr, \diste) &= \frac{1}{2} \int_{\mathcal{V}_2} \sqrt{\distr(v)\cdot \diste(v)} \sum_{y\in \mathcal{Y} }| \distr(y|v) - \diste(y|v) | dv
\end{align}
where $y$ to be discrete, and $\mathcal{V}_1$ and $\mathcal{V}_2$ are partitions of $V$ as follows,
\begin{align}
    \mathcal{V}_1 := \{ v \, | \,  \distr(v) \cdot \diste(v) = 0, v \in V \} \\
    \mathcal{V}_2 := \{ v \, | \,  \distr(v) \cdot \diste(v) \neq 0, v \in V \}
\end{align}
We also introduce a comprehensive notation table (as shown in \cref{tab:symbol}), which defines all key variables and metrics used throughout the paper to help understand.
The nontrivial shifts lead to large differences across environments, rendering KD vulnerable to overfitting with different poor teaching. By studying these areas, we can better understand student performance and learn how to deal with it. Further benchmarking details will be provided below.

\begin{table}[htb]
\centering
\caption{Key symbols and their descriptions.}
\resizebox{.9\textwidth}{!}{
\begin{tabular}{cp{13cm}}
\toprule
\textbf{Symbol} & \textbf{Description} \\ 
\midrule

$ e $ & Environment index ($ e \in \{1, 2, \dots, K\} $) \\ 
$ X^e, Y^e $& Sample and its label in environment $ e $, respectively\\ 
$ P^e_{XY} $& Joint distribution of $ X $ and $ Y $ in environment $ e $ \\ 
$\distr, \diste$ & Probability function of training /test environment \\ 
$ D_{tr}, D_{te} $ & Training/Test dataset containing samples from $ K $ distinct environments \\ 
$ Se $ & Semantic factor (determines invariant features of $ X $ and $ Y $)\\ 
$ V $ & Variation factor (affects context features of $ X $, e.g., style, noise)\\ 
$ \mathcal{V}_1,\mathcal{V}_2 $ & Subset of $ V $, respectively responsible for diversity shift and correlation shift between the environments. \\ 
$ F_{\textit{div}}, F_{\textit{cor}} $ & Diversity/Correlation shift metric \\ 
\bottomrule
\end{tabular}
}
    \label{tab:symbol}
\end{table}

\section{Benchmarking Details}
We present an evaluation framework aimed at exploring the various KD methods in response to distributional changes. Our framework involves proposing a benchmark and assessing over 30 algorithms that span a wide spectrum of approaches.  Furthermore, we include the classical benchmark datasets from various domains across distribution shits. 
We believe that our study provides the first systematic evaluation of the effectiveness of various distillation techniques for distribution shifts. Specifically, the distillation benchmark is explored by covering the following areas:

\subsection{Knowledge Transfer Algorithms}
We first focus on the KD algorithm and introduce the following transfer methods in the benchmark, which we categorize into three distinct aspects.
(a) \textit{Logit-based knowledge}: Knowledge Distillation (\textbf{KD}). 
(b) \textit{Feature-based knowledge}: \textbf{Fitnet}, Attention Transfer (\textbf{AT}), Factor Transfer (\textbf{FT}), Activation Boundaries (\textbf{AB}), Neuron Selectivity Transfer (\textbf{NST}).
(c) \textit{Relation-based knowledge}: Similarity-Preserving (\textbf{SP}), Probabilistic Knowledge Transfer (\textbf{PKT}), Variational Information Distillation (\textbf{VID}), Relation Knowledge Distillation (\textbf{RKD}), Correlation Congruence (\textbf{CC}), Contrastive Representation Distillation (\textbf{CRD}), \textbf{ReviewKD}, Decoupled Knowledge Distillation (\textbf{DKD}), Multi-Level Logit Distillation (\textbf{LogitKD}), and Logit Standardization in Knowledge Distillation (\textbf{LSKD}).
For further details and complete descriptions of the algorithms used, see \cref{tab:kd-types}. 
\begin{table*}[!tb]\centering
\caption{Summary of Knowledge Distillation Methods. Here is a more elaborate explanation of the distillation methods mentioned in our benchmark. For the weight item $\beta$ of $\losskd$, we typically use the optimal hyperparameter from the original work if it is specified.}
\label{tab:kd-types}
\resizebox{.95\textwidth}{!}{
\begin{tabular}{l|cllcc}
\toprule
\multicolumn{2}{c}{\textbf{Method}}   & \textbf{Full name} & \textbf{Knowledge types}  & $\losskd$ & \textbf{$\beta$} \\ \midrule
\multicolumn{2}{c}{KD~\citep{hinton2015distilling}}   & Knowledge Distillation& Soft label & - & - \\  \midrule
\multirow{5}{*}{\rotatebox{90}{\small Feature-based}} 
& Fitnet~\citep{romero2014fitnets} & Hints for Thin Deep Nets & Feature representation & $\mathcal{L}_2(\cdot)$ & 100   \\
& AT~\citep{zagoruyko2016paying}  & Attention Transfer & Attention Maps & $\mathcal{L}_2(\cdot)$ &1000 \\
& FT~\citep{kim2018paraphrasing}  & Factor Transfer & Factor Matrix & $\mathcal{L}_1(\cdot)$ & 500 \\
& AB~\citep{heo2019knowledge}  & Activation Boundaries & Separating Hyperplane (ReLU) & $\mathcal{L}_2(\cdot)$ & - \\
& NST~\citep{huang2017like} & Neuron Selectivity Transfer & Neuron Selectivity Patterns & $\mathcal{L}_{\text{MMD}}(\cdot)$ & 50 \\ \midrule
\multirow{10}{*}{\rotatebox{90}{\small Relation-based}} 
& SP~\citep{tung2019similarity}     & Similarity-Preserving    & Similarity Matrix & $\| \cdot \|_F$ & 3000  \\
& PKT~\citep{passalis2018learning}    & Probabilistic Knowledge  & Similarity Probability Distribution & $\mathcal{L}_{KL}(\cdot)$ & 30000 \\
& VID~\citep{ahn2019variational} & Variational Information Distillation & Variational Information &$\mathcal{L}_{MI}(\cdot)$ & 1 \\
& RKD~\citep{park2019relational} & Relation Knowledge Distillation & Geometric Relations & $\mathcal{L}_H(\cdot),\mathcal{L}_{AW}(\cdot)$ & 25 \\
& CC~\citep{peng2019correlation}  & Correlation Congruence & Instance Relation  & $\mathcal{L}_2(\cdot)$ & 0.02 \\
& CRD~\citep{tian2019contrastive} & Contrastive Representation Distillation & Contrastive Representation & $\mathcal{L}_{KL}(\cdot)$ & 0.8 \\ 
& ReviewKD~\citep{Chen2021Review} & Distilling Knowledge via Knowledge Review & Cross-level Features &$\mathcal{L}_2(\cdot)$ & 8 \\ 
& DKD~\citep{Zhao2022Decoupled} & Decoupled Knowledge Distillation & Target / Non-target Information &$\mathcal{L}_2(\cdot)$ & 1 \\
& LogitKD~\citep{jin2023logit} & Multi-Level Logit Distillation & Multi-level Logit Alignment & $\mathcal{L}_2(\cdot)$& 0.01 \\
 &  LSKD~\citep{sun2024logit} & Logit Standardization in KD & Logit Standardization &$\mathcal{L}_2(\cdot)$ & 0.01 \\
\bottomrule[1pt]
\end{tabular}
}
\end{table*}

\subsection{Data Manipulation Techniques.}
The unseen distribution encourages us to choose suitable manipulation techniques to increase the data quality and diversity.
The manipulation algorithms in our benchmark contain the following areas of specific interest: (1) \textit{vanilla}: \textbf{Identity}. 
(2) \textit{Data Augmentation}: a) \textit{Random-based Augmentation}: \textbf{ImageNet baseline}, \textbf{AutoAugment}~\citep{cubuk2018autoaugment}, \textbf{RandAugment}~\citep{cubuk2020randaugment}, \textbf{Gaussian noise}, b) \textit{Generation-based Augmentation}: \textbf{Mixup}~\citep{zhang2017mixup}, \textbf{CutMix}~\citep{yun2019cutmix}, \textbf{DomainMix}~\citep{wang2020domainmix}, \textbf{MixStyle}~\citep{zhou2021domain}. 
(3) \textit{Data Pruning}: \textbf{Random Prune}, \textbf{EL2N}~\citep{paul2021deep}, \textbf{GraNd}~\citep{paul2021deep}. 
All hyperparameters follow the original settings. 
For further details and comprehensive descriptions of the data techniques used, please refer to \cref{tab:data_manipulation}.
\begin{table*}[tb]
\centering
\caption{Summary of Data Manipulation Techniques.}
\resizebox{.95\textwidth}{!}{
\begin{tabular}{c|cl}
\toprule[1pt]
\multicolumn{2}{c}{\textbf{Method}} & \textbf{Description} \\
\midrule
\multirow{8}{*}{\rotatebox{90}{\footnotesize Augmentation}} 
 & ImageNet baseline & Random cropping and horizontal flipping for image classification. \\
 & AutoAugment & Reinforcement learning to find optimal augmentation policies. \\
 & RandAugment & Simplified search for optimal augmentations with two hyperparameters. \\
 & Gaussian noise & Adds random noise to improve model robustness. \\
 & Mixup & Creates new images and labels by convex combinations. \\
 & CutMix & Cuts and pastes patches from two images and labels. \\
 & DomainMix & Mixes images from different domains. \\
 & MixStyle & Mixes style features of images from different domains. \\
\midrule
\multirow{3}{*}{\rotatebox{90}{\footnotesize Pruning}} 
 & Random Prune & Randomly selects a subset of training data. \\
 & EL2N & Prunes data points with low importance scores based on model prediction. \\
 & GraNd & Prunes data points with low gradient norm difference scores. \\
\bottomrule[1pt]
\end{tabular}
}
\label{tab:data_manipulation}
\end{table*}

\subsection{Optimization Options}
We also investigate the effectiveness of various optimization options on the KD process in addressing the distribution shift. Concretely, we examine a range of optimization options, which include
\textbf{(a) Distillation hyperparameters.}
We perform a sweep for hyperparameters and set for all, which ensures fair comparisons in different settings. 
\textbf{(b) Pretrain or not.} Students pre-training on ImageNet is a prevalent out-of-the-box solution, providing more relevant and diverse features for the target domains, and thus improving the performance.  In our benchmark, we add the pre-training option and observe whether is useful. 
\textbf{(c) Optimizer.} We allow two common optimizers Adam~\citep{kingma2014adam} and SGD~\citep{bottou2012stochastic} in our evaluation, and study the potential shift effects of different optimizers.
\textbf{(d) Student Selection.} 
Learning ability could be specific to student capacity and ResNet-like architectures. In our benchmark, we also replace the student model with another to investigate relevance, such as WRNet and MobileNet. 

\subsubsection{Shifted Datasets}
We choose five datasets for evaluation to explore the performance of KD on the two-dimensional shift of diversity and correlation. 
In terms of diversity shift, we use three widely used benchmark for OOD generalization research: PACS~\citep{li2017deeper}, OfficeHome~\citep{venkateswara2017deep}, and DomainNet~\citep{peng2019moment}. 

\begin{figure*}[tb]
    \centering
    \includegraphics[width=.95\linewidth]{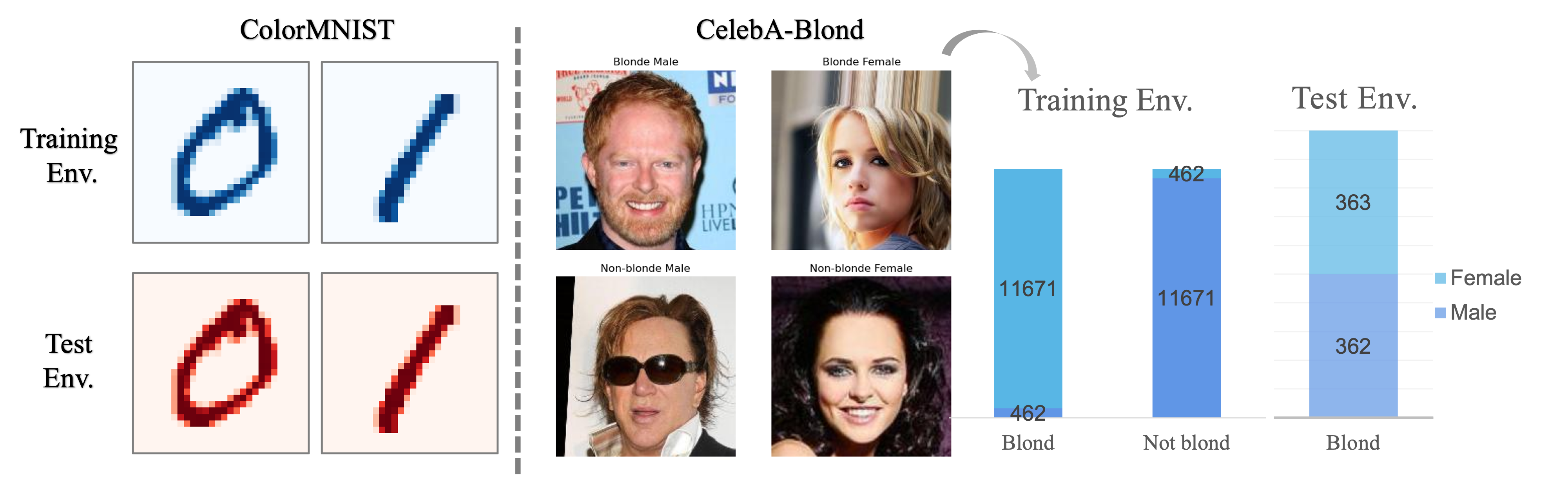}
    \caption{Example inputs for correlation-shifted datasets in our benchmark. \textit{(Left) ColorMNIST:} A modified MNIST dataset where each digit class is associated with a specific color, introducing a spurious correlation between color and digit identity (target label).  \textit{(Right) CelebA-Blond:}  A subset of the CelebA dataset where blond hair is correlated with gender (target label) in training set but not in the test set.}
    \label{fig:cor_dataset}
\end{figure*}

Also, most image datasets are not produced for distribution shift, but one modifies them with artificial transformations to simulate such shifts, such as ColorMNIST and CelebA-Blond~\citep{arjovsky2019invariant}. The variant of original datasets with well-designed makes it available for further study and verification on correlation shift. 
We also visualize some examples in \cref{fig:cor_dataset}.
(1) ColorMNIST is a modified version of MNIST handwritten digit classification dataset. In this variant, the label is a noise-correlated function of the digit and color (blue or red) with flip probabilities $\{0.2, 0.1, 0.9\}$. Notably, the conditional probability of color and digit class differs between training and testing, resulting in spurious correlations. The dataset comprises 60,000 examples of dimension $(2, 28, 28)$ and 2 classes.
(2) CelebA-Blond is a widely used dataset in AI fairness and distribution shift research. It comprises over 200,000 images of celebrity faces, each annotated with 40 attributes. In line with previous studies, we selected a subset of 27,040 images and adjusted the joint distribution of gender and blonde attributes to achieve different shift strengths in different environments.
For the constructed datasets, the training and test domains are predefined and fixed. 
It is worth noting that most of the OOD datasets can be used for our benchmarking, including but not limited to the above five datasets.  

\begin{table}[tb]    
    \centering
    \caption{The details of five popular datasets for distribution shift. To study and mitigate the effects of distribution shift, researchers often rely on specific datasets that are known to exhibit this behavior. Here, we discuss five popular datasets that are commonly used to simulate distribution shifts.}
    \begin{tabular}{lcccc}
    \toprule
         Dataset&  \#Domains& \#Catagories&  \#Examples& Image Type\\ \midrule
         PACS&  4&  7&  10k& Mixed\\
         OfficeHome&  4&  65&  15k& Mixed\\
         DomainNet&  6&  345&  58k& Mixed\\
         ColorMNIST&  3&  2& - & Digits\\
         CelebA-Blond&  3&  2&  50k& Face\\ \bottomrule
    \end{tabular}
    \label{tab:dataset}
\end{table}

\subsection{Evaluation Implementation}
The following implementation choices are highlighted to achieve a consistent and realistic assessment setting.
We select ResNet-50 as the base teacher model, which is a common choice for previous KD and distribution shift algorithms~\citep{wiles2021fine, gou2021knowledge}.
The teacher model is pre-trained in ImageNet and then distilled for each dataset.
We using the same model selection criterion as the out-of-distribution generalization community, and there is still no consensus recently. We choose Training-Domain Validation as a criterion for consistency with existing work~\citep{gulrajani2020search}. 

\subsection{Evaluation Metrics}
In our benchmarks, we include two metrics used as different aspects of the evaluation criteria. Along with the gold standard, \emph{Average accuracy} and \emph{Worst-Group Accuracy} (WGA).
Average accuracy refers to the average accuracy of the model across all environments, and worst-group accuracy refers to the lowest accuracy among all environments of the model. we also report \emph{Expected Calibration Error} (ECE) as an evaluation criterion for the robustness of student models. 
The selected criteria of accuracy were intentionally chosen to carefully evaluate the student model in both average and worst-case scenarios. By carefully considering such cases, we can develop a more comprehensive understanding of its capabilities and limitations, allowing us to make informed decisions about its use. We also provide insight into how the student model performs on different dimensions by investigating the relationship between calibration and accuracy. This analysis can help us identify where there may be room for improvement in KD process and how to balance the accuracy and reliability of predictions.
\section{A Fine-grained Analysis}
\label{sec:analysis}

\begin{table*}[!htb] 
\centering
\caption{\textbf{Average accuracy} for all KD algorithms on datasets dominated by diversity/correlation shift (Teacher/Student: ResNet-50/18). These experiments compare popular algorithms in five benchmarks under identical conditions. $\Delta \uparrow$ denotes the improvement over vanilla KD. The average and standard deviation were reported based on three seeds. }
\label{tab:main1}
\resizebox{.95\textwidth}{!}{
\begin{tabular}{c|c|ccc|cc|cr}
\toprule[1pt]

\multicolumn{2}{c|}{\textbf{Method}}   & \textbf{PACS} & \textbf{OfficeHome} & \textbf{DomainNet} & \textbf{CelebA-Blond} & \textbf{ColorMNIST} & \textbf{Avg.} & \multicolumn{1}{c}{\textbf{$\Delta \uparrow$}} \\ \midrule

\multicolumn{2}{c|}{Teacher}  & 82.59 & 70.74 & 38.80 & 84.74 & 11.93 & 57.8 & \\
\multicolumn{2}{c|}{w/o KD}  & 75.99 & 63.47 & 34.42 & 83.91 & 11.01 & 53.8 & \\ 
\multicolumn{2}{c|}{KD}           & 81.12  & 65.44 & 37.50  & 84.55  & 12.86  & 56.3 & \multicolumn{1}{c}{-} \\ \midrule
\multirow{5}{*}{\rotatebox{90}{Feature-based}}  
& Fitnet      & 73.20 $\pm$ 0.90 & 60.51 $\pm$ 0.80 & 23.86 $\pm$ 2.47 & 85.82 $\pm$ 1.63 & 11.65 $\pm$ 1.15 & 51.0 & -9.41\%               \\
& AT & 80.72 $\pm$ 0.49 & 65.51 $\pm$ 0.13 & 37.39 $\pm$ 0.07 & 84.58 $\pm$ 0.08 & 10.73 $\pm$ 0.66 & 55.8 & -0.89\%               \\
& FT & 79.40 $\pm$ 0.32 & 62.96 $\pm$ 0.30 & 35.76 $\pm$ 0.03 & 84.90 $\pm$ 1.12 & 10.54 $\pm$ 0.27 & 54.7 & -2.84\%               \\
& AB & 76.51 $\pm$ 0.48 & 54.83 $\pm$ 1.10 & 29.98 $\pm$ 0.21 & 85.57 $\pm$ 1.22 &  9.92 $\pm$ 0.15 & 51.4 & -8.70\%               \\
& NST & 82.05 $\pm$ 0.32 & 65.65 $\pm$ 0.29 & 38.05 $\pm$ 0.06 & 84.81 $\pm$ 0.18 & 10.69 $\pm$ 1.35 & 56.2 & -0.18\%               \\ \midrule
\multirow{10}{*}{\rotatebox{90}{Relation-based}} 
& SP & 81.59 $\pm$ 0.51 & 65.20 $\pm$ 0.04 & 34.67 $\pm$ 0.14 & 84.58 $\pm$ 2.21 & 11.05 $\pm$ 0.67 & 55.2 & -1.95\%               \\
& PKT & 81.47 $\pm$ 0.58 & 65.72 $\pm$ 0.17 & 37.93 $\pm$ 0.09 & 84.46 $\pm$ 1.83 & 12.15 $\pm$ 1.29 & 56.3 & \multicolumn{1}{c}{-}               \\
& VID & 80.45 $\pm$ 0.91 & 65.42 $\pm$ 0.20 & 37.55 $\pm$ 0.12 & 85.01 $\pm$ 1.45 & 10.49 $\pm$ 0.32 & 55.8 & -0.89\%               \\
& RKD & 76.11 $\pm$ 0.71 & 46.12 $\pm$ 1.06 & 35.33 $\pm$ 0.15 & 84.37 $\pm$ 0.98 & 10.24 $\pm$ 0.13 & 50.4 & -10.48\%              \\
& CC & 80.42 $\pm$ 1.00 & 65.30 $\pm$ 0.02  & 37.14 $\pm$ 0.08 & 85.43 $\pm$ 0.35 & 12.15 $\pm$ 1.20 & 56.1 & -0.36\%               \\
& CRD & 79.66 $\pm$ 0.15 & 63.91 $\pm$ 0.40 & 37.75 $\pm$ 0.04 & 83.59 $\pm$ 1.82 & 11.48 $\pm$ 1.28 & 55.3 & -1.78\%               \\ 
& ReviewKD & 80.33 $\pm$ 0.40 & 65.74 $\pm$ 0.33 & 37.55 $\pm$ 0.18 & 84.53 $\pm$ 0.97&12.32 $\pm$ 0.57 & 56.1 & -0.36\% \\
& DKD & 81.00 $\pm$ 0.54 & 65.91 $\pm$ 0.32 & 37.17 $\pm$ 0.03 & 84.81 $\pm$ 0.76 & 11.63 $\pm$ 0.34 & 56.1 & -0.36\%\\
& LogitKD & 80.61 $\pm$ 0.67 & 65.97 $\pm$ 0.22 & 39.38 $\pm$ 0.13 & 84.37 $\pm$ 1.19 & 11.14 $\pm$ 0.44 & 56.3 & \multicolumn{1}{c}{-}\\
& LSKD & 78.91 $\pm$ 1.05 & 64.85 $\pm$ 0.28 & 35.79 $\pm$ 0.07 & 84.83 $\pm$ 0.35 & 13.01 $\pm$ 2.50 & 55.5 & -1.42\%
\\

\bottomrule[1pt]
\end{tabular}
}
\end{table*}
\begin{figure*}[!tb]
    \centering
    \subfloat{ \includegraphics[width=0.98\linewidth]{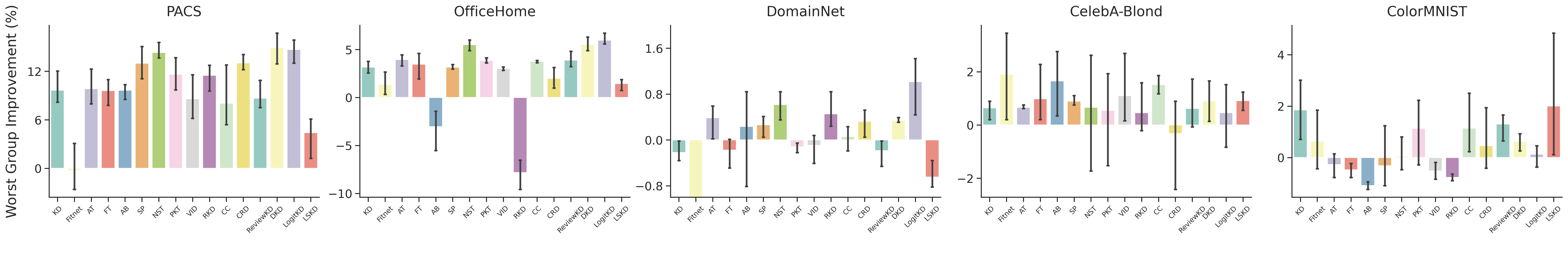}}
    
    \subfloat{ \includegraphics[width=0.98\linewidth]{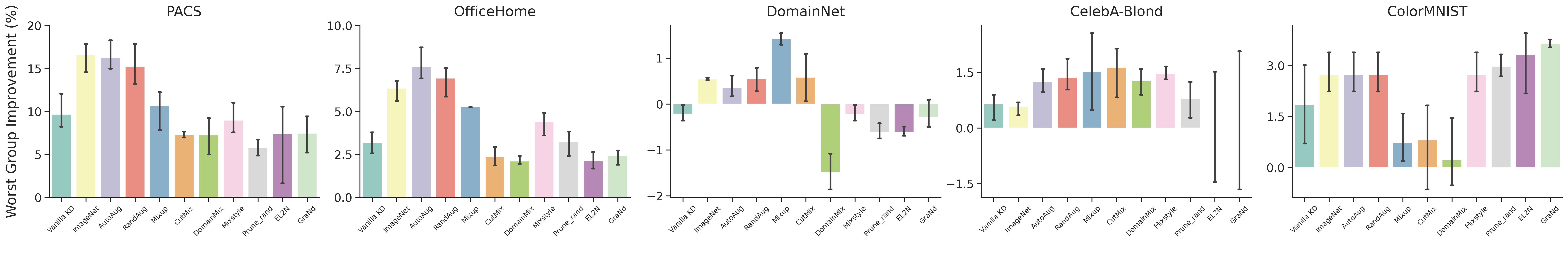}}
    \caption{\textbf{Worst-group accuracy} improvements over ERM on different shifted datasets. \textit{Upper:}Knowledge Transfer algorithms; \textit{Bottom:} data manipulation algorithms. Different algorithms can achieve enhancement only on certain shift types.}
    \label{fig:wga}
\end{figure*}

\subsection{RQ1: Performance Across Distillation Algorithms}
\begin{figure*}[tb]
    \centering
    \includegraphics[width=\linewidth]{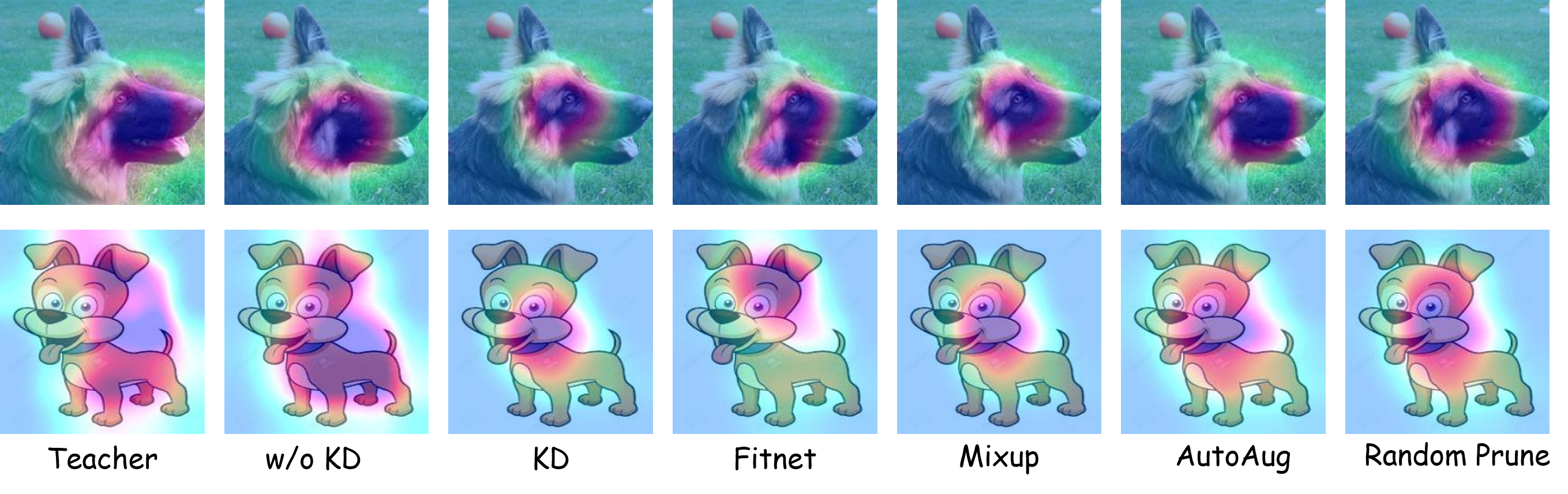}
    \caption{The visualization of Grad-CAM from different models using dogs. Models after distillation pay more attention to the region of interest for specific characteristics. KD can alter the focus of the student, but it is not always useful, such as Fitnet.}
    \label{fig:cam}
\end{figure*}

\paragraph{KD can help lightweight models alleviate the shift effect, but not consistently} 
As \cref{fig:wga} (left) illustrates, we find that all KD algorithms can be effective in improving the worst-group accuracy of the lightweight model. We perform class activation map (CAM)~\citep{selvaraju2017grad} for validity analysis.
It was observed in \cref{fig:cam} that the change in focus towards more general features can enhance the generalization ability of student models. This is primarily due to the simplification of student models, which are often designed to be less complex than their teacher counterparts.  This design choice leads to a preference for learning more general features.  When faced with distribution shifts, these general features may prove more adaptable and generalizable than the intricate features learned by the teacher model. Furthermore, the student model's ability to select and focus on more discriminative features during the learning process plays a crucial role.  This feature selection acts as a form of noise filtering, particularly in the identification of critical regions.  By emphasizing these essential features, the student model can better concentrate on those that are most relevant to the classification task, thereby improving its overall performance, especially in scenarios with distribution shifts.

However, we also observe that no KD method consistently outperforms the average accuracy in both directions of shift. For instance, while improvements are observed for diversity shift in \cref{tab:main1}, none of the methods can effectively identify images on ColorMNIST with a strong correlation shift.  
Based on this inconsistent performance, we can make the following assumpution that although the generalization error of students is improved through distillation, it is limited by the degree of distribution shift and the prediction error of the teacher model. We explore this hypothesis in the following with more experiments.

\paragraph{More complex algorithms offer limited improvement}
Complex algorithms (\vs \textit{vanilla}) exhibit strong performance on the worst-group case, as illustrated in \cref{fig:wga} (left). However, by further zooming in \cref{tab:main1}, we reveal that most methods do not outperform the \emph{vanilla} KD in terms of average accuracy across the most general and challenging cases.
For example, PKT and DKD perform as well as or better than the vanilla KD method on the majority of datasets, while Fitnet underperforms on most datasets. Taking PACS dataset as an example, we can see that: the accuracy of KD is 81.12\%, and NST is relatively better. This may be attributed to the fact that NST forces the student model to match the distributions of neuron selectivity patterns of teachers, which likely provided an advantage in capturing the diversity of the PACS dataset. By doing so, it encouraged the student to learn a richer distribution of features. Overall, different types of KD methods perform differently on different datasets, and complex KD methods offer limits help than vanilla. These findings prompt us to further analyze and understand the underlying reasons for these discrepancies in performance.

\begin{table}[!tb]
\centering
\caption{Results of ResNeXt-50$\times$4 as the teacher.}
\label{tab:r50x}
\resizebox{.95\textwidth}{!}{%
\begin{tabular}{lccccccc}
\toprule
\multirow{2}{*}{\textbf{Method}} & \multirow{2}{*}{\textbf{CelelA-Blond}} &  \multicolumn{5}{c}{\textbf{PACS}} \\ \cmidrule(l){3-7}  
 &  &  \textbf{Art} & \textbf{Cartoon} & \textbf{Photo} & \textbf{Sketch} & \textbf{Avg} \\ 
 \midrule
Teacher & 84.32 & 84.77 & 78.88 & 98.08 & 71.32 & 83.3 \\
KD & 85.64 $\pm$ 0.30 & 80.16 $\pm$ 0.43 & 73.22 $\pm$ 0.67 & 95.25 $\pm$ 0.15 & 66.46 $\pm$ 1.73 & 78.8 \\
AT & 85.38 $\pm$ 0.95  & 79.09 $\pm$ 0.48 & 74.33 $\pm$ 1.29 & 95.29 $\pm$ 0.34 & 68.04 $\pm$ 1.95 & 79.2 \\
DKD & 84.87 $\pm$ 0.32 & 80.47 $\pm$ 1.49 & 76.07 $\pm$ 0.61 & 94.33 $\pm$ 0.19 & 71.21 $\pm$ 0.87 & 80.5 \\
LogitKD & 84.26 $\pm$ 0.70 & 79.62 $\pm$ 1.13 & 74.90 $\pm$ 0.64 & 93.79 $\pm$ 0.28 & 72.79 $\pm$ 1.61 & 80.3 \\ \bottomrule
\end{tabular}%
}
\end{table}

\begin{table}[!tb]
\centering
\caption{Results on PACS for transformer-based models (Teacher: DeiT-Small).}
\label{tab:deit}
\resizebox{.95\textwidth}{!}{%
\begin{tabular}{lcccccc}
\toprule
\textbf{Method} & \textbf{Model} &  \textbf{Art} & \textbf{Cartoon} & \textbf{Photo} & \textbf{Sketch}& \textbf{Avg} \\ \midrule
Teacher & DeiT-Small & 86.82 & 78.24 & 98.68 & 61.01 & 81.2 \\
 \midrule
\multirow{4}{*}{KD} & ConvNeXT-Tiny & 66.03 $\pm$ 16.93 & 59.71 $\pm$ 16.67 & 86.44 $\pm$ 5.63 & 62.55 $\pm$ 27.26 & 68.7 \\
 & DeiT-Tiny & 75.60 $\pm$ 0.02 & 73.15 $\pm$ 1.62 & 94.53 $\pm$ 0.30 & 57.67 $\pm$ 2.44 & 75.2 \\
 & MobileViT-Small & 78.99 $\pm$ 0.54 & 69.07 $\pm$ 2.67 & 95.11 $\pm$ 0.23 & 59.10 $\pm$ 2.98 & 75.6 \\
 & ViT-Tiny & 70.43 $\pm$ 1.74 & 66.89 $\pm$ 2.99 & 92.08 $\pm$ 1.02 & 25.42 $\pm$ 1.82 & 63.7 \\
 \midrule
\multirow{4}{*}{LSKD} & ConvNeXT-Tiny & 21.49 $\pm$ 5.16 & 35.08 $\pm$ 4.05 & 52.85 $\pm$ 13.23 & 32.97 $\pm$ 1.98 & 35.6 \\
 & DeiT-Tiny & 77.10 $\pm$ 0.69 & 71.99 $\pm$ 0.26 & 94.45 $\pm$ 0.14 & 58.81 $\pm$ 2.37 & 75.6 \\
 & MobileViT-Small & 79.67 $\pm$ 1.01 & 69.67 $\pm$ 1.00 & 95.35 $\pm$ 1.05 & 57.68 $\pm$ 5.62 & 75.6 \\
 & ViT-Tiny & 23.49 $\pm$ 4.36 & 65.87 $\pm$ 1.87 & 43.59 $\pm$ 33.10 & 18.68 $\pm$ 0.76 & 37.9 \\
\bottomrule
\end{tabular}%
}
\end{table}

\begin{table}[!tb]
\centering%
 \caption{Results of different layers on $\losskd$ (\%). L1  $\sim$ L4 represents distillation using the knowledge from the first to fourth hint layer.}
 \label{tab:hint}
    \begin{tabular}{lccccc}
    \toprule
    \textbf{Method}  & \textbf{Art}  & \textbf{Cartoon} & \textbf{Photo} & \textbf{Sketch} & \textbf{Avg.} \\ \midrule
    KD     & 81.5 & 78.1    & 95.4  & 69.4  & 81.1 \\
    ~w/ L1 & 74.5 & 75.3    & 91.0  & 62.0   & 75.7 \\
    ~w/ L2 & 70.3 & 72.9    & 89.0  & 54.7   & 71.7 \\
    ~w/ L3 & 67.4 & 71.2    & 88.9  & 61.5   & 72.3 \\
    ~w/ L4 & 81.5 & 78.2    & 94.5  & 74.2   & 82.1 \\ \bottomrule
    \end{tabular}
\end{table}

\paragraph{Larger teacher also offers limited improvement}
The issue remains evident in even when utilizing ResNeXt-50$\times$4 or Transformer-based model, some more powerful architecture, as the teacher model.
Specifically, when employing ResNeXt-50×4 as teacher model under correlation shifts, all KD methods surpassed the teacher model’s performance in  \cref{tab:r50x} . This suggests that the distillation process may effectively suppress spurious feature dependencies learned by the teacher. However, under the diversity shift, all KD methods do not perform significantly better than ResNet-50 as teacher, despite the teacher model achieving an average accuracy of 83.3\%.  
In Transformer-based students, as shown in \cref{tab:deit}, most advanced methods fail to surpass the \emph{vanilla} KD baseline – DeiT-Tiny achieves 75.2\% with KD vs. 75.6\% using LNKD, while ViT-Tiny suffers catastrophic degradation from 63.7\% (KD) to 37.9\% (LNKD).
Distillation effectiveness also exhibits strong architectural compatibility requirements: LNKD improves MobileViT-Small (hybrid CNN-Transformer) by +0.68\% in Art domain but catastrophically fails on pure CNN students like ConvNeXt-Tiny ($68.7\% \rightarrow 35.6\%$).
These observations reveal a dual effect of increasing teacher model capacity, also underscores the necessity of dynamically adjusting distillation strategies based on the specific type of distribution shift encountered.
Besides, the extensibility of our benchmarking framework is demonstrated through systematic evaluation of different architecture pairs (transformers/CNNs/hybrids), providing users with critical guidance for method selection across KD model families.

\begin{figure}[tb]
    \centering
    \includegraphics[width=\linewidth]{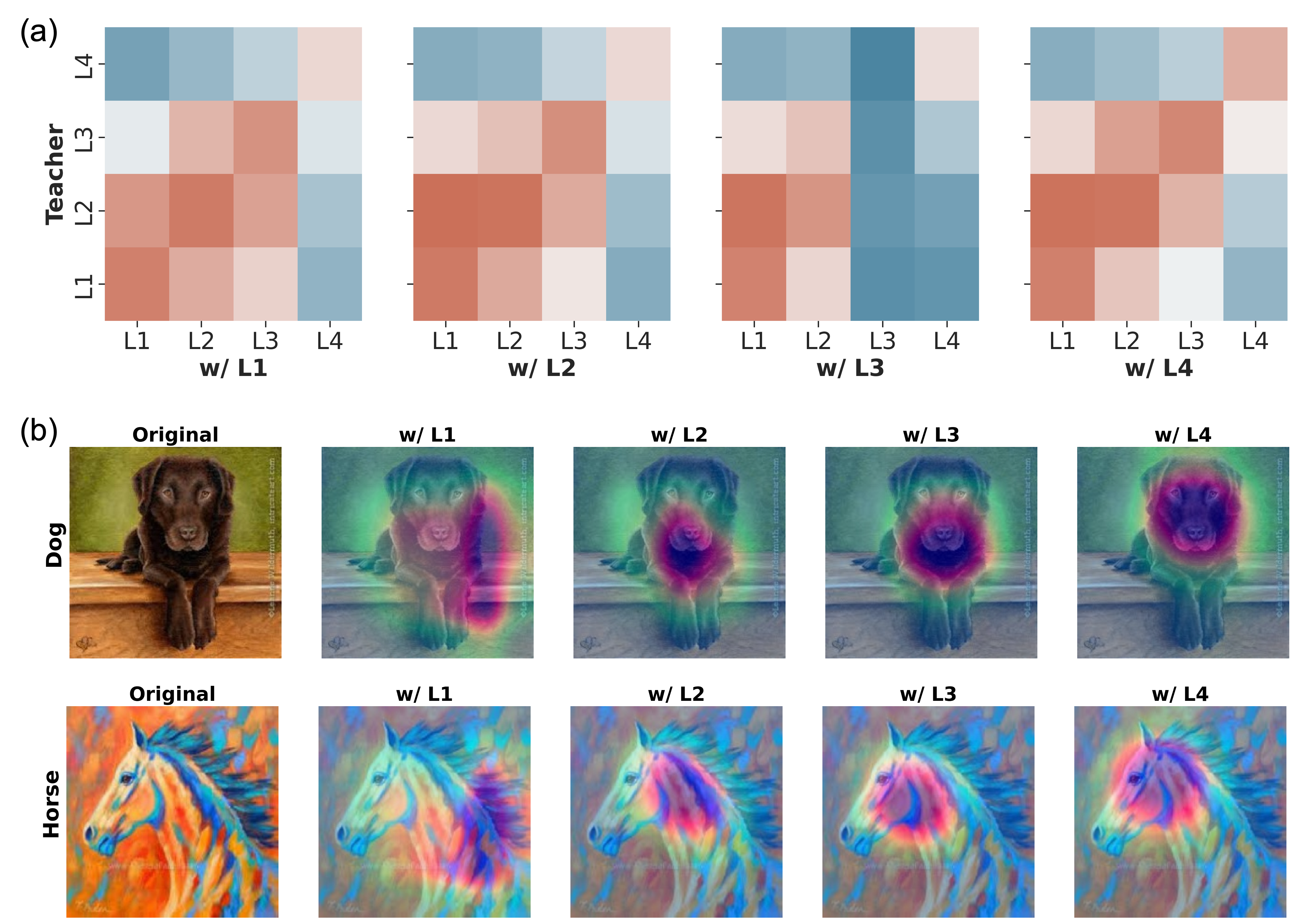}
    \caption{(a) Representation similarity heatmap measured by CKA. L1 $\sim$ L4 represents the representation extracted from the first to fourth hint layer. The CKA inside student differs significantly based on the different layers of knowledge they distilled. (b) The visualization of Grad-CAM from different models using different layers. }
    \label{fig:hint}
\end{figure}

\begin{figure}[tb]
    \centering
    \includegraphics[width=\linewidth]{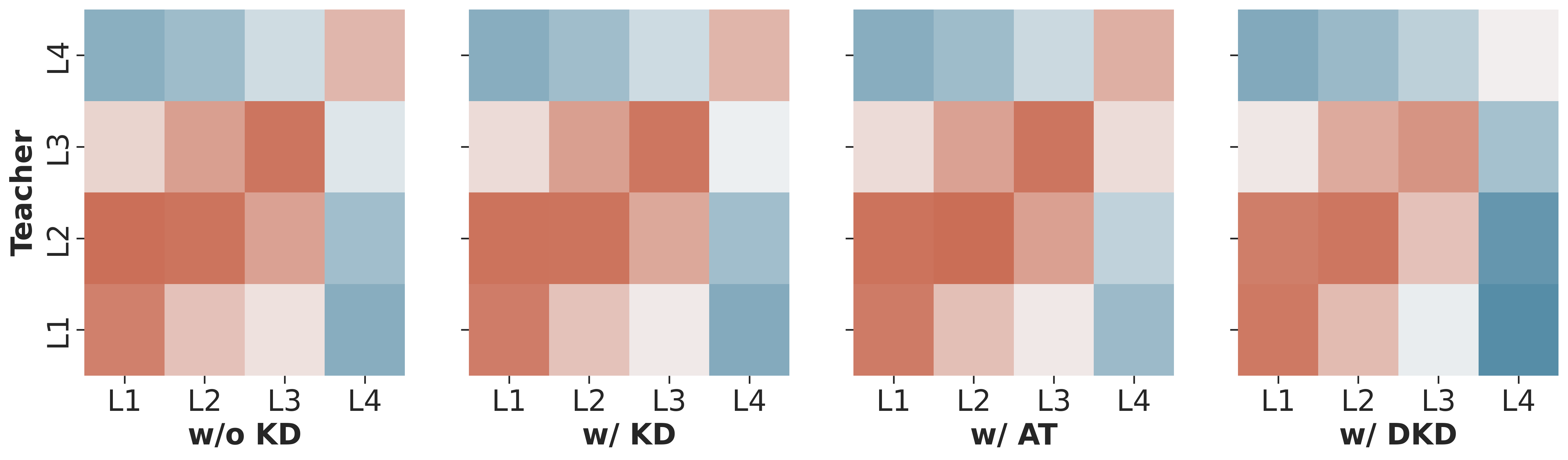}
    \caption{Representation similarity heatmap measured by CKA. L1  $\sim$ L4 represents the representation extracted from the first to fourth hint layer. The CKA inside student differs significantly based on their knowledge sources.}
    \label{fig:cka}
\end{figure}

\paragraph{Low-level knowledge misleads the student while distribution shifts}
To analyze the influence of different knowledge, We first study the similarity of knowledge between different layers through center kernel analysis (CKA)~\citep{kornblith2019similarity}.
As shown in \cref{fig:cka},  the CKA of Fitnet highlights the great diversity between the teacher and student, and its CAM also appeared the same problem. To analyze the influence of knowledge in different hint layers, we further conduct ablation experiments on four variants of the distillation loss term. Feature maps learned by different layers of the teacher model often differed from each other. Unlike~\citep{romero2014fitnets}, which tended to choose the middle layer of the teacher model as the guiding layer, we observe an obvious performance drop after matching the first to the third layer. As reported in \cref{tab:hint}, we find that the last layer achieves the best.

Based on results from CKAs in \cref{fig:hint}(a) and \cref{tab:hint}, the possible reasons for the degradation of complex KD methods can be better explained by the mismatch between transferred features and task-relevant representations. Specifically, over-reliance on unreliable low-level features from the teacher model—especially under distribution shifts—can lead to performance drops. For example, distillation at L3 results in a sharp drop in CKA similarity, indicating severe feature misalignment and suggesting that the transferred knowledge at this layer may interfere with downstream learning.
In contrast, high-level semantic distillation from deeper layers (e.g., L4) consistently improves both feature similarity and classification performance.
High-level semantics can guide students to focus on more important areas, such as the dog face, as shown in \cref{fig:hint}(b).
These results highlight that the student model should not indiscriminately follow the teacher, particularly at non-critical layers, but instead selectively align with domain-independent, semantically meaningful representations to generalize better under domain shifts.

\begin{table}[tb]
\centering
\caption{IRM Penalty Effectiveness on ColorMNIST and OfficeHome dataset. (\%)}
\label{tab:irm}
\resizebox{.95\textwidth}{!}{%
\begin{tabular}{ccccccccc}
\toprule
\multirow{2}{*}{\textbf{Method}} & \multirow{2}{*}{\textbf{Pretrained}} & \multirow{2}{*}{\textbf{Distillation}} & \multirow{2}{*}{\textbf{ColorMNIST}} & \multicolumn{5}{c}{\textbf{OfficeHome}} \\ \cmidrule(l){5-9} 
 &  &  &  & \textbf{Art} & \multicolumn{1}{c}{\textbf{Clipart}} & \multicolumn{1}{c}{\textbf{Product}} & \multicolumn{1}{c}{\textbf{Real World}} & \textbf{Avg.} \\ \midrule
\multirow{2}{*}{Teacher} 
& ERM & -- & 23.77 & 66.79 & 54.36 & 79.21 & 80.56 & 70.23 \\ 
& IRM & -- & 26.58 & 67.04 & 54.32 & 78.33 & 81.48 & 70.29 \\
\midrule

\multirow{2}{*}{w/o KD} 
& ERM & -- & 10.23 & 58.34 & 47.74 & 72.38 & 75.35 & 63.45  \\ 
& IRM & -- & 10.91 & 59.62 & 47.86 & 72.97 & 74.82 & 63.82 \\
\midrule

\multirow{4}{*}{KD}
& ERM & Baseline & 10.28 & 59.70 & 48.85 & 74.32 & 76.11 & 64.75 \\ 
& IRM & Baseline & 11.38 & 60.13 & 48.83 & 74.14 & 76.54 & 64.91 \\
& ERM & + IRM Penalty & 10.24 & 59.90 & 48.83 & 74.32 & 76.04 & 64.77 \\
& IRM & + IRM Penalty & 11.31 & 59.99 & 48.54 & 73.93 & 76.37 & 64.71 \\
\bottomrule
\end{tabular}%
}
\end{table}

\paragraph{Student robustness is fundamentally constrained by both teacher robustness and data biases}

Experimental results in \cref{tab:irm} indicate that the enhancement of OOD robustness in student models through KD significantly depends on the inherent robustness of the teacher model. As shown in \cref{tab:irm}, when using a standard teacher trained with ERM, the student model achieves an accuracy of only 10.28\% on ColorMNIST. In contrast, employing a robust teacher trained with IRM~\citep{arjovsky2019invariant} slightly improves the accuracy to 11.38\%. On the OfficeHome dataset, the average accuracy of the student model guided by the IRM teacher is 64.91\%, compared to 64.75\% with the ERM teacher, reflecting a marginal improvement. Notably, even when incorporating the IRM penalty during the distillation process, the performance difference between students guided by ERM and IRM teachers remains insignificant. These findings reveal an inherent limitation of KD: the student model struggles to surpass the bias boundaries of the teacher model, with its robustness ceiling essentially constrained by the representational capacity of the teacher. When the teacher model exhibits systematic bias, traditional distillation mechanisms tend to perpetuate this bias through KD, hindering fundamental improvements in scenarios involving distribution shifts.

\begin{figure*}[!tb]
    \centering
    \includegraphics[width=\linewidth]{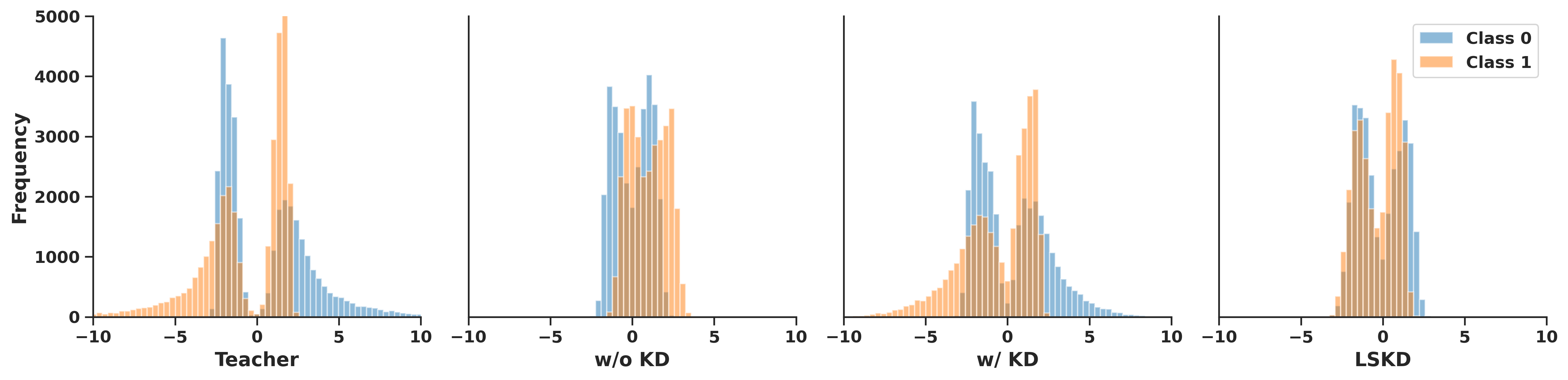}
    \caption{Logit distribution of different methods on test set of ColorMNIST.}
    \label{fig:logit}
\end{figure*}

To gain deeper insights into the significant differences in generalization capabilities observed on ColorMNIST, we investigate the impact of dark knowledge on model performance by analyzing the logit distributions across KD methods. As shown in \cref{fig:logit}, the teacher's outputs exhibit a distinct bimodal structure, indicating high confidence in predictions between two classes. However, without KD, the student model's logit distribution clusters near zero, revealing high uncertaintyr. When vanilla KD is applied, the student's logit distribution partially recovers the teacher's bimodal structure, yet inter-class separation remains low—a limitation also observed with LSKD. Furthermore, \cref{tab:irm} demonstrates that applying the IRM penalty under various settings fails to substantially improve CMNIST performance, regardless of whether KD is employed.  Distillation helps maintain prediction confidence and class separation at the logit level, but its effectiveness is still limited. This further supports the hypothesis that the primary challenge in distribution shifts stems from dual biases in both the teacher and training data, which collectively prevent students from getting robustness.

\paragraph{Extra information is required}
Although the heterogeneity of the data caused by distribution shifts poses challenges to distillation, this heterogeneity is not fully explored during distillation. We can directly leverage additional context information, such as domain indexes, or explicitly group data into contexts (or domains) to help students learn~\citep{arjovsky2019invariant, chen2022pareto}. 
Through the guidance of extra information, students may better observe spurious-correlated features and learn invariant features, even if the teacher is not robust.

\emph{\underline{In brief}}, dark knowledge, which is useful in the case of \iid has the opposite effect under distribution shift, since the representation learned by the teacher model may be context and style information, leading to overfitting.

\subsection{RQ2: The Role of Distillation Data}
\begin{table*}[!tb] \centering
\caption{Average accuracy for all data techniques on datasets dominated by diversity/correlation shift (Teacher/Student: ResNet-50/18). These experiments compare 12 popular algorithms in five benchmarks under identical conditions, showing the robustness and improvement potential of data techniques. $\Delta \uparrow$ denotes the improvement over vanilla KD. The average and standard deviation were reported based on three seeds.}
\label{tab:main2}
\resizebox{.95\textwidth}{!}{
\begin{tabular}{c|l|ccc|cc|cr}
\toprule[1pt]
\multicolumn{2}{c|}{\textbf{Method}}     & \textbf{PACS} & \textbf{OfficeHome} & \textbf{DomainNet} & \textbf{CelebA-Blond} & \textbf{ColorMNIST} & \textbf{Avg.} & \multicolumn{1}{c}{\textbf{$\Delta \uparrow$}} \\ \midrule
\multicolumn{2}{c|}{\textit{vanilla} KD} & 81.12 $\pm$ 0.65 & 65.44 $\pm$ 0.36 & 37.50 $\pm$ 0.02  & 84.55 $\pm$ 0.38 & 12.86 $\pm$ 1.15 & 56.3 & \multicolumn{1}{c}{-} \\ \midrule
\multirow{8}{*}{\rotatebox{90}{Augmentation}}
& + ImageNet    & 83.24 $\pm$ 0.10 & 67.31 $\pm$ 0.22 & 36.80 $\pm$ 0.07  & 84.48 $\pm$ 0.20 & 13.73 $\pm$ 0.60 & 57.1 & 1.45\%                \\
& + AutoAug     & 83.56 $\pm$ 0.28 & 67.44 $\pm$ 0.17 & 36.73 $\pm$ 0.08 & 85.15 $\pm$ 0.32 & 13.73 $\pm$ 0.60 & 57.3 & 1.83\%                \\
& + RandAug     & 83.00 $\pm$ 0.61  & 67.26 $\pm$ 0.30 & 37.11 $\pm$ 0.03 & 85.27 $\pm$ 0.44 & 13.73 $\pm$ 0.60 & 57.3 & 1.74\%                \\
& + Mixup       & 80.07 $\pm$ 0.67 & 65.74 $\pm$ 0.06 & 39.39 $\pm$ 0.09 & 85.43 $\pm$ 1.04 & 11.73 $\pm$ 0.76 & 56.5 & 0.32\%                \\
& + CutMix      & 78.70 $\pm$ 0.37  & 65.07 $\pm$ 0.40 & 38.00 $\pm$ 0.18  & 85.54 $\pm$ 0.71 & 11.83 $\pm$ 1.30 & 55.8 & -0.83\%               \\
& + DomainMix   & 79.61 $\pm$ 0.22 & 65.43 $\pm$ 0.10 & 34.84 $\pm$ 0.13 & 85.17 $\pm$ 0.35 & 11.24 $\pm$ 1.08 & 55.3 & -1.84\%               \\
& + Mixstyle    & 80.63 $\pm$ 0.39 & 65.90 $\pm$ 0.46  & 37.50 $\pm$ 0.02  & 85.38 $\pm$ 0.17 & 13.73 $\pm$ 0.60  & 56.6 & 0.59\%                \\ 
& + Noise       & 66.78 $\pm$ 2.27 & 52.74 $\pm$ 1.69 & 18.84 $\pm$ 0.21 & 76.38 $\pm$ 1.93 & 10.37 $\pm$ 0.72 & 45.0 & -20.02\%              \\ \midrule
\multirow{3}{*}{\rotatebox{90}{Pruning} } 
& + Prune\_rand & 79.43 $\pm$ 0.39 & 64.78 $\pm$ 0.22 & 37.09 $\pm$ 0.01 & 84.69 $\pm$ 0.49 & 13.99 $\pm$ 0.33 & 56.0 & -0.53\%               \\
& + EL2N        & 79.02 $\pm$ 1.69 & 64.20 $\pm$ 0.34  & 36.24 $\pm$ 0.04 & 83.89 $\pm$ 1.49 & 14.33 $\pm$ 1.01 & 55.5 & -1.35\%               \\
& + GraNd       & 79.05 $\pm$ 0.54 & 64.14 $\pm$ 0.04 & 36.82 $\pm$ 0.02 & 83.89 $\pm$ 1.91 & 14.65 $\pm$ 0.12 & 55.7 & -1.04\%               \\ 
\bottomrule[1pt]
\end{tabular}
}
\end{table*}

\paragraph{Data augmentation helps marginally}
It is worth noting that data augmentation is a powerful and effective tool for distribution shift and KD separately, but the effect becomes weaker when combined.
As \cref{fig:wga} (right) reveals, the effectiveness of augmentation depends on the realistic situation of datasets. 
Previous views are that any form of data augmentation can be effective as it increases the likelihood of a student matching across multiple domains. 
However, the different strategy shows diverse results in \cref{tab:main2}. Of all shift settings, random-based augmentation typically improves performance, while generation-based augmentation performs better on correlation shift, such as AutoAugment versus CutMix.  
Regarding the underwhelming performance of Gaussian noise, our analysis suggests that the addition of Gaussian noise can not effectively align the natural distribution of the data or the inherent noise present in the teacher’s knowledge.  This could result in a less effective transfer of knowledge from the teacher to the student model, leading to poorer performance. 

\paragraph{Student learning with random-based augmentation performs better but is context-dependent} 
Generation-based augmentation methods like Mixup and CutMix have proven powerful tools; however, our benchmarks reveal that they underperform compared to random-based augmentations such as RandAugment and AutoAugment. This underperformance can be attributed to several factors related to KD under distribution shifts. 
Firstly, generative augmentation alters the original data distribution by mixing samples, which can introduce biases that impede the student model's ability to generalize. Secondly, effective knowledge transfer relies on aligning the student model's training data with that of the teacher model. Random augmentations facilitate this alignment by enhancing diversity without introducing irrelevant features, thereby maintaining a closer connection to the teacher's learned knowledge. Lastly, random augmentations offer greater stability and adaptability since they are easier to control compared to the hyperparameter-sensitive nature of generative methods.

Moreover, the assumption that random augmentation methods are inherently effective for handling KD under domain shifts may seem self-evident, but it is, in fact, quite complex. While random augmentation can improve performance in our benchmarks, its success heavily depends on the nature of the data and the specific distribution shifts encountered. Additionally, the belief that employing random transformations aligned with real data will automatically enhance student model performance during KD neglects the intricacies of knowledge transfer. The effectiveness of these transformations is contingent on their alignment with invariant features; without empirical validation, proximity to real data may inadvertently introduce noise or misalignment that detracts from performance. Ultimately, the effectiveness of data augmentation in KD is not as straightforward as it appears, emphasizing the necessity for a nuanced understanding of their interactions with distribution shifts and knowledge transfer dynamics, as well as the need for rigorous empirical validation to support these assertions.

\paragraph{Data pruning makes sense for students under distribution shift} 
Our observation in \cref{tab:main2} suggests that not all training samples are equal, with data pruning exploring data quality required for distillation by keeping important examples. Specifically, we observed that selecting 75\% of all samples and utilizing distillation techniques yielded a student model that retained over 95\% of its pre-pruning performance. Moreover, we found that random pruning outperformed other well-designed metrics. 

Interestingly, our results indicate that data pruning even can outperform data augmentation techniques in correlation shift-dominated datasets, such as ColorMNIST. 
The reason for this interesting result may be that it optimizes the quality of training samples and reduces overfitting, and the superiority of random pruning may be because it can avoid the bias that may be introduced by artificially designed metrics, while also being able to effectively reduce noise and redundancy in the training data.
Overall, our study provides empirical evidence highlighting the significance of carefully selecting data for distillation purposes. These insights could inform the creation and utilization of distillation datasets.

\paragraph{High-quality KD data is needed}
The distillation data needed for the student model can be original, or augmentation, or pruning. This depends on the structure and goals of the KD model. Augmenting the data can enhance its diversity and robustness, while pruning can reduce redundancy and noise. 
However, in any case, manipulation aims to make some useful changes to our training data so that we can get more high-quality data and its distribution is closer to the invariant one across any distribution.

\emph{\underline{In short}}, Data augmentation is effective for distribution shift and knowledge distillation separately, but not as effective when combined. Data pruning can be a valuable technique for KD by selecting important examples, even outperforming data augmentation in certain scenarios.

\subsection{RQ3: Possible Causes on Training Options}

In addition to our primary findings, we conducted a thorough examination of key factors that impact the optimization program. Our investigation yielded several intriguing observations, which are as follows.

\begin{table}[tb]
\centering 
\caption{The performance of Pre-training and Optimizer Selection(\%). The training strategy can have a strong impact on the distribution shift during the distillation process.}
 \label{tab:optim}
    \begin{tabular}{lccccc} 
    \toprule
    \textbf{Method}    & \textbf{Art}  & \textbf{Cartoon} & \textbf{Photo} & \textbf{Sketch} & \textbf{Avg.} \\ \midrule
    Baseline     & 81.5 & 78.1    & 95.4  & 69.4  & 81.1 \\
    ~w/~~ Adam         & 51.7 & 62.9    & 70.7  & 61.2 & 61.6  \\
    ~w/o Pretrain & 33.6 & 40.6    & 51.3  & 27.4 & 38.2  \\
    \bottomrule
    \end{tabular}
\end{table}

\paragraph{Pre-training may help, but depends on student tasks}
Pre-training can be a valuable technique for distillation tasks to distribution shift, provided that the features learned during pre-training are beneficial for the student's performance (\cref{tab:optim}). However, it also indicates that if the representations needed for the student tasks differ from those learned during pre-training, the benefit may be significantly limited, especially in PACS different sub-task (\texttt{Photo} versus \texttt{Sketch}).
Meanwhile, if the teacher model has biases on the training dataset, these errors can be transferred to the student model during the KD process (\texttt{Color} versus \texttt{Digit} in ColorMNIST task).

\begin{figure}[!tb]
    \centering
    \subfloat{ \includegraphics[width=.85\linewidth]{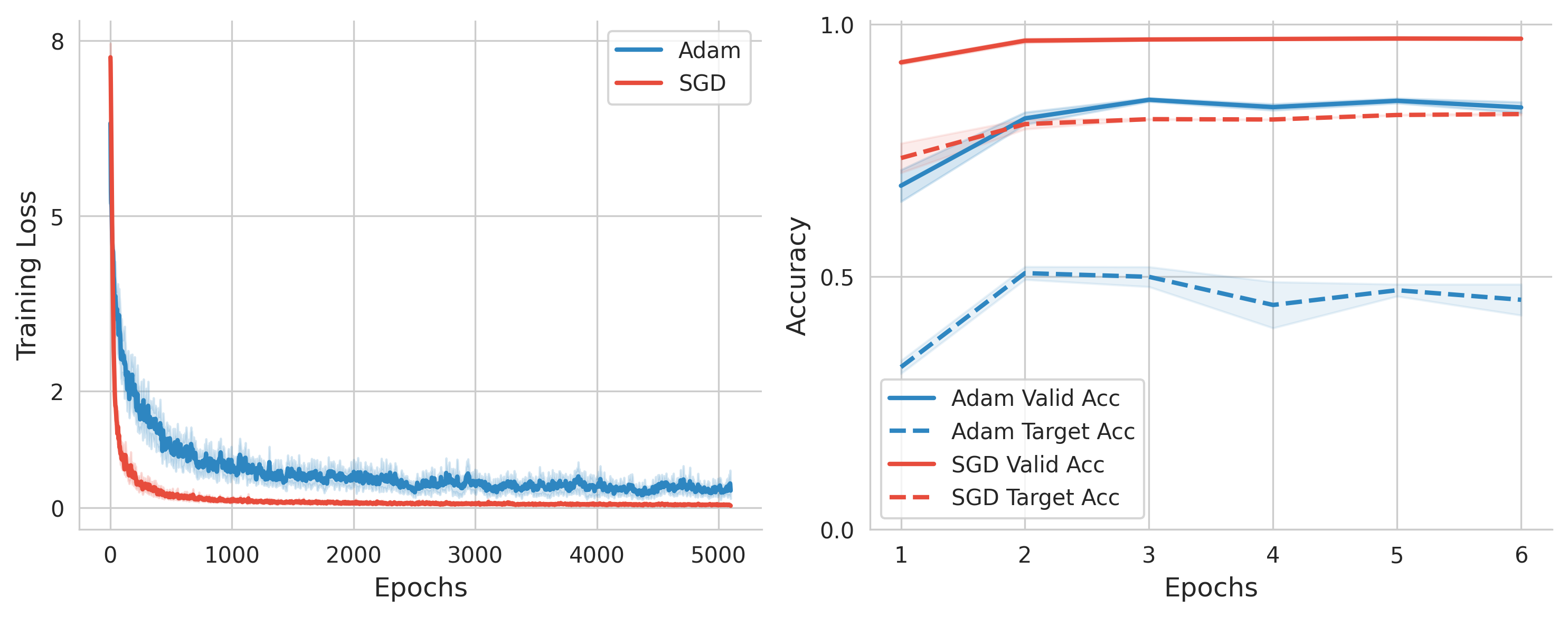} \label{fig:opt_loss}}
    
    \subfloat{ \includegraphics[width=.85\linewidth]{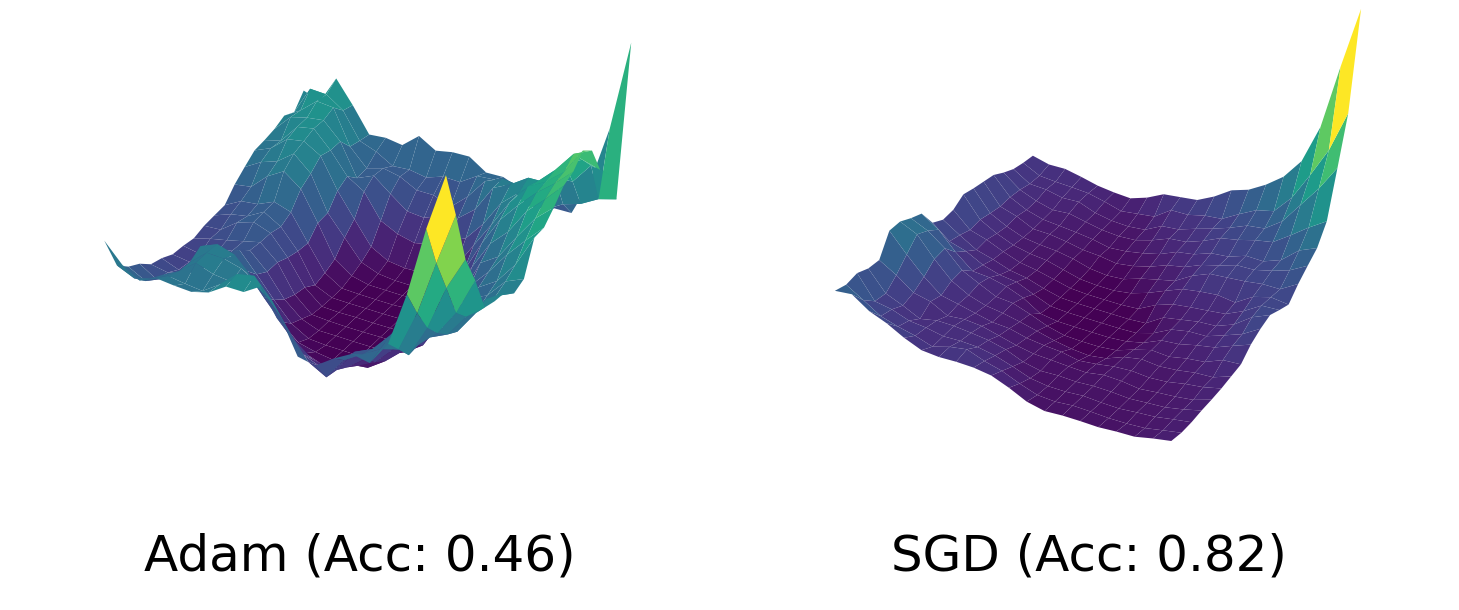} \label{fig:opt_landscape}}
    \caption{(Upper) Training dynamics for different optimizers, including training loss and accuracy. (Bottom) Loss surface after training with different optimizers.}
    \label{fig:opt}
\end{figure}

\paragraph{SGD for KD on distribution shift, not Adam}
While Adam and SGD are typically comparable in their performance in KD, our evaluation revealed a notable deviation from this norm. Specifically, we observed a significant decrease in performance when utilizing Adam compared to SGD in \cref{tab:optim}. This is further observed by the training dynamics in \cref{fig:opt_loss}: while Adam exhibits decent training loss and validation accuracy, its test accuracy suffers severe fluctuations. Through our visualizations of the loss landscape in \cref{fig:opt_landscape}, the stochastic nature of SGD enables it to find escapes from local optima and find flat minima, whereas Adam converges to sharp minima, whose narrow curvature amplifies sensitivity to distribution shifts. Our observation is in line with the results in \citep{zou2023understanding}, highlighting the shift weaknesses of adaptive optimizers, such as Adam.

\paragraph{Results of other student architectures}
We further evaluate our benchmark using other small networks. KD still has an overall boosting effect on tiny models. However, the extent of gain varies depending on the domain and structure. (\cref{tab:model}).
Different student model architectures may differ in their sensitivity to knowledge distillation. For example, due to its smaller number of parameters, MobileNet may not possess the representational capabilities of more complex models like ResNet. Consequently, MobileNet may require additional care and fine-tuning to ensure the effective transfer of knowledge during the KD process.
At the same time, KD can generally improve the performance on Art and Cartoon, but not on Photo or Sketch across different structures.
Some domains, such as Art and Cartoon, may have more distinct features that allow KD to transfer knowledge more effectively. Conversely, in domains like Photo or Sketch, features can be more subtle or complex, making knowledge transfer more difficult across models.
These results suggest that improving the robustness of these models to distribution shift may be challenging, particularly in the worst-case scenario.

\begin{table}[tb]
    \centering
    \caption{Gaps between different student models before/after KD (\%).}
    \label{tab:model}
    \begin{tabular}{lrccccc}
    \toprule
    \textbf{Model}         & \textbf{Params}   & \textbf{Art}  & \textbf{Cartoon} & \textbf{Photo} & \textbf{Sketch} & \textbf{Avg.} \\ \midrule
    ResNet-18     & 11.35M   & 4.5 & 6.1     & 0.2   & 9.7    & 5.1  \\
    ResNet-34     & 21.29M   & 0.5 & 0.7     & 0.3   & 1.8    & 0.8  \\
    WRN-16-2   & 0.70M    & 1.2 & 2.0     & 1.1   & -7.1   & -0.7 \\
    WRN-40-2   & 2.26M    & 3.5 & -0.2    & -1.1  & -2.6   & -0.1 \\
    MobileNet-S & 0.93M & 4.0 & 0.2     & -0.2  & -1.0   & 0.8  \\
    MobileNet-L & 2.98M & 3.2 & 0.8     & 0.6   & 4.3    & 2.2  \\ \bottomrule
    \end{tabular}
\end{table}

\begin{figure*}[!tb]
    \centering
    \includegraphics[width=\linewidth]{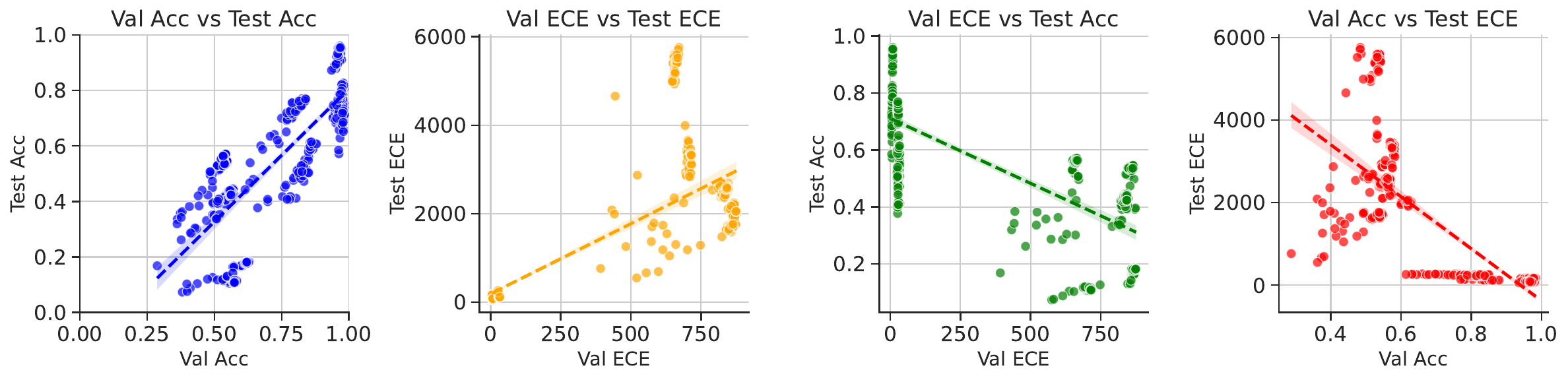}
    \caption{The role of calibration in assessing the robustness of student models under distribution shift. The weak correlation between accuracy and ECE suggests that high accuracy does not guarantee good calibration, emphasizing the need to consider both accuracy and calibration when assessing the robustness of KD algorithms.}
    \label{fig:metric}
\end{figure*}

\paragraph{The role of calibration in student models under distribution shift}
In our analysis, we examined the relationships between various performance metrics of student models under distribution shift, focusing on test accuracy and Expected Calibration Error (ECE). A lower ECE indicates better alignment between predicted confidence and its actual accuracy, reflecting improved calibration. Well-calibrated models are more reliable, especially when encountering data distributions that differ from the training set.
As shown in \cref{fig:metric}, we observed a strong negative correlation between accuracy and ECE values, suggesting that as models become more accurate, their calibration error tends to decrease. This is a desirable outcome, as it implies that improvements in accuracy are accompanied by enhancements in calibration. However, the correlation between test accuracy and validation ECE was weak, indicating that good validation ECE cannot be used as a reliable indicator of test accuracy.
This discrepancy may result from the distribution shift between the validation and test distributions. dditionally, the ECE metric has known limitations, such as sensitivity to binning strategies and potential discontinuities, which can affect its reliability.

In conclusion, calibration is a critical factor in assessing the robustness of KD. Even with high accuracy, poor calibration can lead to suboptimal performance in real-world applications. Therefore, the objective of KD should encompass not only enhancing model accuracy but also improving calibration performance, ensuring consistency between predicted confidence and actual accuracy when facing unfamiliar data distributions.

\subsection{More Discussion}
\label{sec:more_discuss}
\paragraph{Performance across different types of shift}
Several algorithms improve much better over the pre-KD period on the situation of diversity shift. However, in the case of ColorMNIST and CelebA-Blond, most of KD algorithms cannot minimize the gap between teacher and student. Consequently, we argue that previous KD methods remain largely vulnerable to spurious correlations. In particular, vanilla KD still achieves good results, but the problem is that students receive bias while being taught. The use of data manipulation can get rid of bias to some extent, but this is limited by how relevant the manipulated data is to the ground truth.

\paragraph{The selection of distillation methods based solely on the prior is not a viable option}
 Although our evaluation and comparison based on quantitative distribution shifts will be helpful to practitioners to choose a better method. However, in real-world applications, data collection in unknown environments is expensive, so how to precisely pinpoint distribution shifts and deploy lightweight models on end-side is an important area in the future.
 
\paragraph{Our negative claims should be critically considered}
Our negative statements (e.g., ``Method A does not work better than KD against distribution shift") are based on the fact that such statements are setting-specific and context-constrained. However, one of our key contributions is to introduce a new perspective on the distribution shift for knowledge distillation, even in model compression.

\subsection{Connecting KD to Information Theory}
\label{sec:theory}

We now provide some intuition that reinterpret KD  through the lens of information theory and offer new theoretical insights, especially under the context of distribution shift~.
Specifically, $I(X;Y)$ denotes the mutual information (MI) between $X$ and $Y$, which can be given following  Donsker-Varadhan  variational representation~\citep{polyanskiy2014lecture}:
\begin{equation}\label{eq:dv}
    \begin{aligned}
         & I(X; Y) = \sup_{c: \mathcal{X} \times \mathcal{Y} \mapsto \mathbb{R}}\mathcal{I}_c(X; Y), \\
         & \mathcal{I}_c(X; Y) = \mathbb{E}_{x, y \sim P_{XY}} c(x, y) - \log \mathbb{E}_{x, y \sim P_X \times P_Y}(e^{c(x, y)}),
    \end{aligned}
\end{equation}
where $c$ is a critic function\citep{poole2019variational} measuring output alignment.

Traditional KD aims to transfer knowledge from a teacher model to a student by minimizing the KL divergence between their output distributions. With its connection to \emph{information-maximization} principles~\citep{tsai2020self, ahn2019variational}, we have $I(S; T)$ which denotes the MI between the student's and teacher's outputs for input data $X$. However, the (asymptotic) equivalence of  KD objective optimization and  student-teacher MI maximization does not necessarily imply good performance under distribution shift.
Recent progress on information theory~\citep{tsai2020self,li2023s} emphasize the importance of selectively transmitting information, suggesting that context-variant components should be filtered during distillation to improve student robustness in non-\iid settings.
Let $I(S; T \mid Y)$ denote the conditional MI of student $S$ and teacher $T$ given the target $Y$. By applying the chain rule~\citep{polyanskiy2014lecture} of mutual information, we obtain the following identity:
\begin{equation} \label{eq:info_relation}
    \underbrace{I(S; Y)}_{\text{supervised goal}} = \underbrace{I(S; T)}_{\text{KD goal}} + \underbrace{I(S; Y\mid T)}_{\text{Domain-invariant}} - \underbrace{I(S; T\mid Y)}_{\text{Domain-variant}}.
\end{equation} 
In KD process, $ I(S; T) $, as the core objective, measures the efficiency of knowledge transfer from the teacher to the student. Under ideal conditions, $ I(S; Y\mid T) $ represents the student $ S $'s additional learning capacity about the true label $ Y $ given the teacher's knowledge $ T $, while $ I(S; T\mid Y) $ captures redundant information between the student and teacher that is irrelevant to $ Y $.
A small value of $I(S; Y\mid T)$ is  a standard assumption in information-theoretic characterizations of transfer learning~\citep{wu2024generalization, bao2019information}.
However, under distribution shift, the statistical dependence between $ T $ and $ Y $ is disrupted by spurious correlations or style shift, preventing the student from effectively acquiring $ Y $-related information through $ T $. The relation \eqref{eq:info_relation} leads to insufficient distillation from $ I(S; Y\mid T) $ and amplified negative impacts from $ I(S; T\mid Y) $, collectively causing a significant decline in $ I(S;Y) $.
Specifically,
\begin{equation} \label{eq:shift_relation}
    I(S; Y) \uparrow  = \underbrace{I(S; T)}_{\text{Misleading}\uparrow } + \underbrace{I(S; Y\mid T)}_{\text{Compensation} \downarrow } - \underbrace{I(S; T\mid Y)}_{\text{Redundancy}\uparrow}.
\end{equation} 
The relation in \cref{eq:shift_relation} suggests that under distribution shift, $I(S; T)$ may transfer misleading knowledge, $I(S; Y \mid T)$ becomes less informative due to the teacher’s degraded performance, and the redundancy captured by $I(S; T \mid Y)$ tends to increase. Together, these factors contribute to a significant reduction in $I(S; Y)$, thereby weakening the student’s domain-invariant representation.
In extreme cases, if the teacher is almost completely ineffective (there is a strong false correlation between $T$ and $Y$), the student performance may be lower than that of independent training.

Furthermore, the mechanisms behind the failure of existing KD methods under distribution shift can be explained via the decomposition in \cref{eq:info_relation}. In FitNet, the student is encouraged to closely mimic intermediate layers of the teacher. When spurious correlations exist in such layers, the student inherits these through $I(S; T \mid Y)$, which reinforces domain-variant patterns and leads to overfitting. NST matches Gram matrices to align style statistics, where higher-order statistics implicitly capture domain-invariant structures (e.g., object shapes), thus suppressing domain-specific interference by reducing $I(S; T \mid Y)$. Logit-level distillation transfers class probabilities via soft labels, preserving semantic information. However, if the teacher’s logits encode spurious patterns (e.g., bimodal color bias in ColorMNIST), the student still inherits this bias, leaving residual context-variant information in $I(S; T \mid Y)$.

\section{Related works}
Two areas are closely related to our research, namely out-of-distribution generalization and knowledge distillation, and we briefly review previous studies as follows.

\textbf{Knowledge distillation.} 
Large deep neural networks have achieved remarkable success, especially in scenarios with large-scale data. However, deploying large deep models on resource-limited devices, such as mobile phones and edge devices, is challenging because of their computational requirements.
To develop lightweight deep networks, \citet{hinton2015distilling} first proposed the concept of knowledge distillation, which replaces one-hot ground truth with `dark knowledge'' and defined the teaching manner. This paradigm allows for easier and more flexible deployment of models in edge devices and has been improved and proven efficient in a wide range of practical applicants.
Several follow-up studies~\citep{GAO2024106587,LAN2025107133} extend vanilla KD by redesigning knowledge types to guide the learning styles of student models. Such methods can be categorized into two types, feature-based methods and relation-based methods. Feature-based distillation methods used primarily hidden layer representation, \ie, activation, and feature map~\citep{kim2018paraphrasing, heo2019knowledge}. Relation-based distillation methods focused more on the useful relationships between different neurons, layers, or sample pairs~\citep{tung2019similarity, tian2019contrastive}. 
For example,  MTKD-SSR~\citep{gou2023multi} enhances KD by integrating stage-wise distillation and cross-stage review,  enabling the student model to self-reflection for improved knowledge absorption.
Besides, FFKD~\citep{gou2024reciprocal} introduces a bidirectional distillation framework where student feedback dynamically guides teacher refinement.
However, their focus was primarily on generalization under the assumption of \iid.
In contrast, we are concerned with the robustness of KD when this assumption is violated due to distribution shifts.

\textbf{Benchmarking distribution shift.}
Big data often comes from multiple domains, with distribution shifts across sources, leading to a significant failure of model generalization. To tackle the problem of distribution shifts when training with multiple domains and generalizing to unseen domains, existing methods can be divided into three groups:
(1) Data augmentation.  \citep{wang2020domainmix} enhanced the generalization performance of OoD models by improving the quantity and diversity of data from multiple domains. 
(2) Representation learning. Representation learning can perform inter-domain feature alignment via kernels~\citep{hu2020domain}, adversarial learning~\citep{ganin2015unsupervised}, \etc or minimize the invariant risk to learn domain-invariant features~\citep{arjovsky2019invariant} or disentangle features into domain-shared or specific parts for better generalization~\citep{zhang2022towards}. 
(3) Learning strategy. This kind of method enhanced the generalization ability with the usage of general learning strategies, such as ensemble learning~\citep{li2022domain} and meta learning~\citep{li2018learning}.
Recent studies have discovered that ERM methods outperformed state-of-the-art on average when carefully tuned and implemented~\citep{gulrajani2020search}.
Differently from previous studies, we investigate when and how distribution shifts can be mitigated through distillation learning. Through our thorough and systematic studies, we observe that KD can effectively enhance the robustness of small models in unknown environments, especially for the worst group.

\textbf{Benchmarking knowledge distillation}  
To systematically evaluate and compare different KD approaches, several benchmark studies have been conducted across various tasks and domains. 
For instance, \citet{tian2019contrastive} evaluated 12 KD methods on CIFAR-100 and ImageNet, demonstrating the effectiveness of proposed CRD.
In NLP, \citet{yuan2024impact} assessed the energy efficiency and performance of KD by benchmarking BERT-like models across three tasks. For ranking models, RD-Suite~\citep{qin2023rd} provides a comprehensive evaluation framework over four real-world datasets, covering both standard and transfer distillation settings.
Despite these efforts, limited attention has been paid to the behavior of KD under distribution shift. Existing works focus primarily on improving KD performance in standard settings~\citep{fang2021mosaicking,zhou2022device,do2022momentum}. For example, \citet{fang2021mosaicking} leveraged out-of-domain data to improve in-domain distillation, \citet{zhou2022device} enhanced out-of-distribution (OOD) generalization through disruptive data augmentation, and \citet{do2022momentum} proposed a momentum-based strategy for data-free KD. However, these studies do not explicitly address the non-\iid nature of real-world deployments.
No studies have discussed how knowledge distillation methods work under non-\iid situations.
We explore and discuss how KD methods work without \iid assumption. It is a new setting for KD and is the practical setting in the real world.  We propose a systematic evaluation framework to benchmark knowledge distillation against distribution shifts.

\section{Conclusion}
\textbf{Top Takeaways.}
Distribution shift resulted in poor KD performance, so there must be no free lunch when choosing KD methods for lightweight students. According to various distribution shift goals and our evaluation results, but also in response to the research questions in \cref{sec:intro},  here are some simple takeaways to help stakeholders consider.

\begin{itemize}
    \item \textbf{Knowledge distillation works.} 
    Our benchmarks show that KD is crucial for lightweight models prone to overfitting under distribution shifts.   Therefore, KD is still recommended when someone wants to obtain robust lightweight models.
    
    \item \textbf{Exploring novel algorithms is required.} Complex KD methods show limited improvement over vanilla KD across distribution shifts. While vanilla KD partially handle the shift, making other methods robust to shifts remains a challenge, especially for the correlation shift. 
    
    \item \textbf{Augmentation cannot guarantee robust student.} Data augmentation can aid KD in extracting general features, but it’s not always reliable. Exploring new data-level methods for enhancing student model robustness is a promising direction under distribution shifts.
    
    \item \textbf{Do not forget pre-training and SGD.} Pre-training and the choice of optimizer are important considerations for KD under distribution shifts, warranting further study on training strategies.
\end{itemize}

\textbf{Conclusion.}
We formulate the novel KD paradigm, \textsc{ShiftKD}, under distribution shifts and broaden the learning objectives of knowledge distillation to multiple domains to address real application scenarios. We propose a systematic evaluation framework from three diverse perspectives including the knowledge distillation algorithms, data manipulation mechanisms, and optimization options, and we take a comprehensive evaluation benchmark covering more than 30 methods for five benchmark datasets. 
Several novel tips based on our benchmark are summarized to allow the research community to find optimal solutions when applying knowledge distillation techniques in real scenarios.

\section*{Acknowledgement}
This work was supported in part by Guangdong Basic and Applied Basic Research Foundation (2023A1515012848), and CCF-DiDi GAIA Collaborative Research Funds for Young Scholars, and Science and Technology Research Program of Chongqing Municipal Education Commission, China (Grant No. KJZD-K202400703).

\section*{CRediT authorship contribution statement}
\textbf{Songming Zhang:} Conceptualization, Methodology, Software, Writing-Original draft preparation. 
\textbf{Yuxiao Luo:} Software, Writing-Original draft preparation. 
\textbf{Ziyu Lyu:} Supervision, Writing-Reviewing and Editing.
\textbf{Xiaofeng Chen:} Supervision, Writing-Reviewing and Editing.

\bibliographystyle{plainnat} 
\bibliography{paper}

\begin{thebibliography}{63}
\providecommand{\natexlab}[1]{#1}
\providecommand{\url}[1]{\texttt{#1}}
\expandafter\ifx\csname urlstyle\endcsname\relax
  \providecommand{\doi}[1]{doi: #1}\else
  \providecommand{\doi}{doi: \begingroup \urlstyle{rm}\Url}\fi

\bibitem[Ahn et~al.(2019)Ahn, Hu, Damianou, Lawrence, and Dai]{ahn2019variational}
Sungsoo Ahn, Shell~Xu Hu, Andreas Damianou, Neil~D Lawrence, and Zhenwen Dai.
\newblock Variational information distillation for knowledge transfer.
\newblock In \emph{Proceedings of the IEEE/CVF Conference on Computer Vision and Pattern Recognition}, pages 9163--9171, 2019.

\bibitem[Arjovsky et~al.(2019)Arjovsky, Bottou, Gulrajani, and Lopez-Paz]{arjovsky2019invariant}
Martin Arjovsky, L{\'e}on Bottou, Ishaan Gulrajani, and David Lopez-Paz.
\newblock Invariant risk minimization.
\newblock \emph{arXiv preprint arXiv:1907.02893}, 2019.

\bibitem[Bao et~al.(2019)Bao, Li, Huang, Zhang, Zheng, Zamir, and Guibas]{bao2019information}
Yajie Bao, Yang Li, Shao-Lun Huang, Lin Zhang, Lizhong Zheng, Amir Zamir, and Leonidas Guibas.
\newblock An information-theoretic approach to transferability in task transfer learning.
\newblock In \emph{2019 IEEE international conference on image processing (ICIP)}, pages 2309--2313. IEEE, 2019.

\bibitem[Bottou(2012)]{bottou2012stochastic}
L{\'e}on Bottou.
\newblock Stochastic gradient descent tricks.
\newblock In \emph{Neural Networks: Tricks of the Trade: Second Edition}, pages 421--436. Springer, 2012.

\bibitem[Chen et~al.(2021)Chen, Liu, Zhao, and Jia]{Chen2021Review}
Pengguang Chen, Shu Liu, Hengshuang Zhao, and Jiaya Jia.
\newblock Distilling knowledge via knowledge review.
\newblock In \emph{Proceedings of the IEEE/CVF Conference on Computer Vision and Pattern Recognition (CVPR)}, pages 5008--5017, June 2021.

\bibitem[Chen et~al.(2022)Chen, Zhou, Bian, Xie, Ma, Zhang, Yang, Han, and Cheng]{chen2022pareto}
Yongqiang Chen, Kaiwen Zhou, Yatao Bian, Binghui Xie, Kaili Ma, Yonggang Zhang, Han Yang, Bo~Han, and James Cheng.
\newblock Pareto invariant risk minimization.
\newblock \emph{arXiv preprint arXiv:2206.07766}, 2022.

\bibitem[Cubuk et~al.(2018)Cubuk, Zoph, Mane, Vasudevan, and Le]{cubuk2018autoaugment}
Ekin~D Cubuk, Barret Zoph, Dandelion Mane, Vijay Vasudevan, and Quoc~V Le.
\newblock Autoaugment: Learning augmentation policies from data.
\newblock \emph{arXiv preprint arXiv:1805.09501}, 2018.

\bibitem[Cubuk et~al.(2020)Cubuk, Zoph, Shlens, and Le]{cubuk2020randaugment}
Ekin~D Cubuk, Barret Zoph, Jonathon Shlens, and Quoc~V Le.
\newblock Randaugment: Practical automated data augmentation with a reduced search space.
\newblock In \emph{Proceedings of the IEEE/CVF conference on computer vision and pattern recognition workshops}, pages 702--703, 2020.

\bibitem[Do et~al.(2022)Do, Le, Nguyen, Nguyen, Harikumar, Tran, Rana, and Venkatesh]{do2022momentum}
Kien Do, Thai~Hung Le, Dung Nguyen, Dang Nguyen, Haripriya Harikumar, Truyen Tran, Santu Rana, and Svetha Venkatesh.
\newblock Momentum adversarial distillation: Handling large distribution shifts in data-free knowledge distillation.
\newblock \emph{Advances in Neural Information Processing Systems}, 35:\penalty0 10055--10067, 2022.

\bibitem[Fang et~al.(2021)Fang, Bao, Song, Wang, Xie, Shen, and Song]{fang2021mosaicking}
Gongfan Fang, Yifan Bao, Jie Song, Xinchao Wang, Donglin Xie, Chengchao Shen, and Mingli Song.
\newblock Mosaicking to distill: {{Knowledge}} distillation from out-of-domain data.
\newblock In \emph{Advances in {{Neural Information Processing Systems}}}, volume~34, pages 11920--11932, 2021.

\bibitem[Ganin and Lempitsky(2015)]{ganin2015unsupervised}
Yaroslav Ganin and Victor Lempitsky.
\newblock Unsupervised domain adaptation by backpropagation.
\newblock In \emph{International conference on machine learning}, pages 1180--1189. PMLR, 2015.

\bibitem[Gao et~al.(2024)Gao, Shi, Hu, Feng, Zhu, Wan, and Feng]{GAO2024106587}
Liqing Gao, Peng Shi, Lianyu Hu, Jichao Feng, Lei Zhu, Liang Wan, and Wei Feng.
\newblock Cross-modal knowledge distillation for continuous sign language recognition.
\newblock \emph{Neural Networks}, 179:\penalty0 106587, 2024.

\bibitem[Gou et~al.(2021)Gou, Yu, Maybank, and Tao]{gou2021knowledge}
Jianping Gou, Baosheng Yu, Stephen~J Maybank, and Dacheng Tao.
\newblock Knowledge distillation: A survey.
\newblock \emph{International Journal of Computer Vision}, 129:\penalty0 1789--1819, 2021.

\bibitem[Gou et~al.(2023)Gou, Xiong, Yu, Du, Zhan, and Tao]{gou2023multi}
Jianping Gou, Xiangshuo Xiong, Baosheng Yu, Lan Du, Yibing Zhan, and Dacheng Tao.
\newblock Multi-target knowledge distillation via student self-reflection.
\newblock \emph{International Journal of Computer Vision}, 131\penalty0 (7):\penalty0 1857--1874, 2023.

\bibitem[Gou et~al.(2024)Gou, Chen, Yu, Liu, Du, Wan, and Yi]{gou2024reciprocal}
Jianping Gou, Yu~Chen, Baosheng Yu, Jinhua Liu, Lan Du, Shaohua Wan, and Zhang Yi.
\newblock Reciprocal teacher-student learning via forward and feedback knowledge distillation.
\newblock \emph{IEEE transactions on multimedia}, 2024.

\bibitem[Gulrajani and Lopez-Paz(2020)]{gulrajani2020search}
Ishaan Gulrajani and David Lopez-Paz.
\newblock In search of lost domain generalization.
\newblock \emph{arXiv preprint arXiv:2007.01434}, 2020.

\bibitem[Heo et~al.(2019)Heo, Lee, Yun, and Choi]{heo2019knowledge}
Byeongho Heo, Minsik Lee, Sangdoo Yun, and Jin~Young Choi.
\newblock Knowledge transfer via distillation of activation boundaries formed by hidden neurons.
\newblock In \emph{Proceedings of the AAAI Conference on Artificial Intelligence}, volume~33, pages 3779--3787, 2019.

\bibitem[Hinton et~al.(2015)Hinton, Vinyals, and Dean]{hinton2015distilling}
Geoffrey Hinton, Oriol Vinyals, and Jeff Dean.
\newblock Distilling the knowledge in a neural network.
\newblock \emph{arXiv preprint arXiv:1503.02531}, 2015.

\bibitem[Hu et~al.(2020)Hu, Zhang, Chen, and Chan]{hu2020domain}
Shoubo Hu, Kun Zhang, Zhitang Chen, and Laiwan Chan.
\newblock Domain generalization via multidomain discriminant analysis.
\newblock In \emph{Uncertainty in Artificial Intelligence}, pages 292--302. PMLR, 2020.

\bibitem[Huang and Wang(2017)]{huang2017like}
Zehao Huang and Naiyan Wang.
\newblock Like what you like: Knowledge distill via neuron selectivity transfer.
\newblock \emph{arXiv preprint arXiv:1707.01219}, 2017.

\bibitem[Jin et~al.(2023)Jin, Wang, and Lin]{jin2023logit}
Ying Jin, Jiaqi Wang, and Dahua Lin.
\newblock Multi-level logit distillation.
\newblock In \emph{Proceedings of the IEEE/CVF Conference on Computer Vision and Pattern Recognition (CVPR)}, pages 24276--24285, June 2023.

\bibitem[Kim et~al.(2018)Kim, Park, and Kwak]{kim2018paraphrasing}
Jangho Kim, SeongUk Park, and Nojun Kwak.
\newblock Paraphrasing complex network: Network compression via factor transfer.
\newblock \emph{Advances in neural information processing systems}, 31, 2018.

\bibitem[Kingma and Ba(2014)]{kingma2014adam}
Diederik~P Kingma and Jimmy Ba.
\newblock Adam: A method for stochastic optimization.
\newblock \emph{arXiv preprint arXiv:1412.6980}, 2014.

\bibitem[Kornblith et~al.(2019)Kornblith, Norouzi, Lee, and Hinton]{kornblith2019similarity}
Simon Kornblith, Mohammad Norouzi, Honglak Lee, and Geoffrey Hinton.
\newblock Similarity of neural network representations revisited.
\newblock In \emph{International conference on machine learning}, pages 3519--3529. PMLR, 2019.

\bibitem[Lan et~al.(2025)Lan, Li, Yan, Xiang, Tang, Wu, and Chen]{LAN2025107133}
Zhen Lan, Zixing Li, Chao Yan, Xiaojia Xiang, Dengqing Tang, Min Wu, and Zhenghua Chen.
\newblock Rmkd: Relaxed matching knowledge distillation for short-length ssvep-based brain–computer interfaces.
\newblock \emph{Neural Networks}, 185:\penalty0 107133, 2025.
\newblock ISSN 0893-6080.

\bibitem[Le et~al.(2025)Le, Nguyen, Thwal, Qiao, Zhang, and Hong]{LE2025107017}
Huy~Q. Le, Minh~N.H. Nguyen, Chu~Myaet Thwal, Yu~Qiao, Chaoning Zhang, and Choong~Seon Hong.
\newblock Fedmekt: Distillation-based embedding knowledge transfer for multimodal federated learning.
\newblock \emph{Neural Networks}, 183:\penalty0 107017, 2025.
\newblock ISSN 0893-6080.

\bibitem[Li et~al.(2017)Li, Yang, Song, and Hospedales]{li2017deeper}
Da~Li, Yongxin Yang, Yi-Zhe Song, and Timothy~M Hospedales.
\newblock Deeper, broader and artier domain generalization.
\newblock In \emph{Proceedings of the IEEE international conference on computer vision}, pages 5542--5550, 2017.

\bibitem[Li et~al.(2018)Li, Yang, Song, and Hospedales]{li2018learning}
Da~Li, Yongxin Yang, Yi-Zhe Song, and Timothy Hospedales.
\newblock Learning to generalize: Meta-learning for domain generalization.
\newblock In \emph{Proceedings of the AAAI conference on artificial intelligence}, volume~32, 2018.

\bibitem[Li et~al.(2023)Li, Wu, Sun, Chen, Tian, Zhu, Meng, Zheng, and Wang]{li2023s}
Jintang Li, Ruofan Wu, Wangbin Sun, Liang Chen, Sheng Tian, Liang Zhu, Changhua Meng, Zibin Zheng, and Weiqiang Wang.
\newblock What's behind the mask: Understanding masked graph modeling for graph autoencoders.
\newblock In \emph{Proceedings of the 29th ACM SIGKDD Conference on Knowledge Discovery and Data Mining}, pages 1268--1279, 2023.

\bibitem[Li et~al.(2022)Li, Ren, Jiang, Li, Zhang, and Li]{li2022domain}
Ziyue Li, Kan Ren, Xinyang Jiang, Bo~Li, Haipeng Zhang, and Dongsheng Li.
\newblock Domain generalization using pretrained models without fine-tuning.
\newblock \emph{arXiv preprint arXiv:2203.04600}, 2022.

\bibitem[Park et~al.(2019)Park, Kim, Lu, and Cho]{park2019relational}
Wonpyo Park, Dongju Kim, Yan Lu, and Minsu Cho.
\newblock Relational knowledge distillation.
\newblock In \emph{Proceedings of the IEEE/CVF Conference on Computer Vision and Pattern Recognition}, pages 3967--3976, 2019.

\bibitem[Passalis and Tefas(2018)]{passalis2018learning}
Nikolaos Passalis and Anastasios Tefas.
\newblock Learning deep representations with probabilistic knowledge transfer.
\newblock In \emph{Proceedings of the European Conference on Computer Vision (ECCV)}, pages 268--284, 2018.

\bibitem[Paul et~al.(2021)Paul, Ganguli, and Dziugaite]{paul2021deep}
Mansheej Paul, Surya Ganguli, and Gintare~Karolina Dziugaite.
\newblock Deep learning on a data diet: Finding important examples early in training.
\newblock \emph{Advances in Neural Information Processing Systems}, 34:\penalty0 20596--20607, 2021.

\bibitem[Peng et~al.(2019{\natexlab{a}})Peng, Jin, Liu, Li, Wu, Liu, Zhou, and Zhang]{peng2019correlation}
Baoyun Peng, Xiao Jin, Jiaheng Liu, Dongsheng Li, Yichao Wu, Yu~Liu, Shunfeng Zhou, and Zhaoning Zhang.
\newblock Correlation congruence for knowledge distillation.
\newblock In \emph{Proceedings of the IEEE/CVF International Conference on Computer Vision}, pages 5007--5016, 2019{\natexlab{a}}.

\bibitem[Peng et~al.(2019{\natexlab{b}})Peng, Bai, Xia, Huang, Saenko, and Wang]{peng2019moment}
Xingchao Peng, Qinxun Bai, Xide Xia, Zijun Huang, Kate Saenko, and Bo~Wang.
\newblock Moment matching for multi-source domain adaptation.
\newblock In \emph{Proceedings of the IEEE/CVF international conference on computer vision}, pages 1406--1415, 2019{\natexlab{b}}.

\bibitem[Polyanskiy and Wu(2014)]{polyanskiy2014lecture}
Yury Polyanskiy and Yihong Wu.
\newblock Lecture notes on information theory.
\newblock \emph{Lecture Notes for ECE563 (UIUC) and}, 6\penalty0 (2012-2016):\penalty0 7, 2014.

\bibitem[Poole et~al.(2019)Poole, Ozair, Van Den~Oord, Alemi, and Tucker]{poole2019variational}
Ben Poole, Sherjil Ozair, Aaron Van Den~Oord, Alex Alemi, and George Tucker.
\newblock On variational bounds of mutual information.
\newblock In \emph{International conference on machine learning}, pages 5171--5180. PMLR, 2019.

\bibitem[Qin et~al.(2023)Qin, Jagerman, Pasumarthi, Zhuang, Zhang, Bai, Hui, Yan, and Wang]{qin2023rd}
Zhen Qin, Rolf Jagerman, Rama~Kumar Pasumarthi, Honglei Zhuang, He~Zhang, Aijun Bai, Kai Hui, Le~Yan, and Xuanhui Wang.
\newblock Rd-suite: A benchmark for ranking distillation.
\newblock \emph{Advances in Neural Information Processing Systems}, 36, 2023.

\bibitem[Romero et~al.(2014)Romero, Ballas, Kahou, Chassang, Gatta, and Bengio]{romero2014fitnets}
Adriana Romero, Nicolas Ballas, Samira~Ebrahimi Kahou, Antoine Chassang, Carlo Gatta, and Yoshua Bengio.
\newblock Fitnets: Hints for thin deep nets.
\newblock \emph{arXiv preprint arXiv:1412.6550}, 2014.

\bibitem[Selvaraju et~al.(2017)Selvaraju, Cogswell, Das, Vedantam, Parikh, and Batra]{selvaraju2017grad}
Ramprasaath~R Selvaraju, Michael Cogswell, Abhishek Das, Ramakrishna Vedantam, Devi Parikh, and Dhruv Batra.
\newblock Grad-cam: Visual explanations from deep networks via gradient-based localization.
\newblock In \emph{Proceedings of the IEEE international conference on computer vision}, pages 618--626, 2017.

\bibitem[Sorscher et~al.(2022)Sorscher, Geirhos, Shekhar, Ganguli, and Morcos]{sorscher2022beyond}
Ben Sorscher, Robert Geirhos, Shashank Shekhar, Surya Ganguli, and Ari Morcos.
\newblock Beyond neural scaling laws: beating power law scaling via data pruning.
\newblock \emph{Advances in Neural Information Processing Systems}, 35:\penalty0 19523--19536, 2022.

\bibitem[Stanton et~al.(2021)Stanton, Izmailov, Kirichenko, Alemi, and Wilson]{stanton2021Does}
Samuel Stanton, Pavel Izmailov, Polina Kirichenko, Alexander~A Alemi, and Andrew~G Wilson.
\newblock Does knowledge distillation really work?
\newblock In \emph{Advances in {{Neural Information Processing Systems}}}, volume~34, pages 6906--6919. {Curran Associates, Inc.}, 2021.

\bibitem[Sun et~al.(2024)Sun, Ren, Li, Wang, and Cao]{sun2024logit}
Shangquan Sun, Wenqi Ren, Jingzhi Li, Rui Wang, and Xiaochun Cao.
\newblock Logit standardization in knowledge distillation.
\newblock In \emph{Proceedings of the IEEE/CVF conference on computer vision and pattern recognition}, pages 15731--15740, 2024.

\bibitem[Tian et~al.(2024)Tian, Xu, and Li]{TIAN2024106567}
Yingjie Tian, Shaokai Xu, and Muyang Li.
\newblock Decoupled graph knowledge distillation: A general logits-based method for learning mlps on graphs.
\newblock \emph{Neural Networks}, 179:\penalty0 106567, 2024.
\newblock ISSN 0893-6080.

\bibitem[Tian et~al.(2019)Tian, Krishnan, and Isola]{tian2019contrastive}
Yonglong Tian, Dilip Krishnan, and Phillip Isola.
\newblock Contrastive representation distillation.
\newblock \emph{arXiv preprint arXiv:1910.10699}, 2019.

\bibitem[Tsai et~al.(2020)Tsai, Wu, Salakhutdinov, and Morency]{tsai2020self}
Yao-Hung~Hubert Tsai, Yue Wu, Ruslan Salakhutdinov, and Louis-Philippe Morency.
\newblock Self-supervised learning from a multi-view perspective.
\newblock \emph{arXiv preprint arXiv:2006.05576}, 2020.

\bibitem[Tung and Mori(2019)]{tung2019similarity}
Frederick Tung and Greg Mori.
\newblock Similarity-preserving knowledge distillation.
\newblock In \emph{Proceedings of the IEEE/CVF International Conference on Computer Vision}, pages 1365--1374, 2019.

\bibitem[Venkateswara et~al.(2017)Venkateswara, Eusebio, Chakraborty, and Panchanathan]{venkateswara2017deep}
Hemanth Venkateswara, Jose Eusebio, Shayok Chakraborty, and Sethuraman Panchanathan.
\newblock Deep hashing network for unsupervised domain adaptation.
\newblock In \emph{Proceedings of the IEEE conference on computer vision and pattern recognition}, pages 5018--5027, 2017.

\bibitem[Wang et~al.(2021{\natexlab{a}})Wang, Liao, Zhao, Cui, and Shao]{wang2020domainmix}
Wenhao Wang, Shengcai Liao, Fang Zhao, Kangkang Cui, and Ling Shao.
\newblock Domainmix: Learning generalizable person re-identification without human annotations.
\newblock In \emph{British Machine Vision Conference}, 2021{\natexlab{a}}.

\bibitem[Wang et~al.(2021{\natexlab{b}})Wang, Li, Chau, and Kot]{wang2021embracing}
Yufei Wang, Haoliang Li, Lap-pui Chau, and Alex~C Kot.
\newblock Embracing the dark knowledge: Domain generalization using regularized knowledge distillation.
\newblock In \emph{Proceedings of the 29th ACM International Conference on Multimedia}, pages 2595--2604, 2021{\natexlab{b}}.

\bibitem[Wiles et~al.(2021)Wiles, Gowal, Stimberg, Rebuffi, Ktena, Dvijotham, and Cemgil]{wiles2021fine}
Olivia Wiles, Sven Gowal, Florian Stimberg, Sylvestre-Alvise Rebuffi, Ira Ktena, Krishnamurthy~Dj Dvijotham, and Ali~Taylan Cemgil.
\newblock A fine-grained analysis on distribution shift.
\newblock In \emph{International Conference on Learning Representations}, 2021.

\bibitem[Wu et~al.(2024)Wu, Manton, Aickelin, and Zhu]{wu2024generalization}
Xuetong Wu, Jonathan~H Manton, Uwe Aickelin, and Jingge Zhu.
\newblock On the generalization for transfer learning: An information-theoretic analysis.
\newblock \emph{IEEE Transactions on Information Theory}, 2024.

\bibitem[Yang et~al.(2023)Yang, Zhang, Katabi, and Ghassemi]{yang2023change}
Yuzhe Yang, Haoran Zhang, Dina Katabi, and Marzyeh Ghassemi.
\newblock Change is hard: A closer look at subpopulation shift.
\newblock In \emph{International Conference on Machine Learning}, 2023.

\bibitem[Ye et~al.(2022)Ye, Li, Bai, Yu, Hong, Zhou, Li, and Zhu]{ye2022ood}
Nanyang Ye, Kaican Li, Haoyue Bai, Runpeng Yu, Lanqing Hong, Fengwei Zhou, Zhenguo Li, and Jun Zhu.
\newblock Ood-bench: Quantifying and understanding two dimensions of out-of-distribution generalization.
\newblock In \emph{Proceedings of the IEEE/CVF Conference on Computer Vision and Pattern Recognition}, pages 7947--7958, 2022.

\bibitem[Yuan et~al.(2024)Yuan, Shi, Zhang, Chen, Zhang, Stoico, and Malavolta]{yuan2024impact}
Ye~Yuan, Jiacheng Shi, Zongyao Zhang, Kaiwei Chen, Jingzhi Zhang, Vincenzo Stoico, and Ivano Malavolta.
\newblock The impact of knowledge distillation on the energy consumption and runtime efficiency of nlp models.
\newblock In \emph{Proceedings of the IEEE/ACM 3rd International Conference on AI Engineering-Software Engineering for AI}, pages 129--133, 2024.

\bibitem[Yun et~al.(2019)Yun, Han, Oh, Chun, Choe, and Yoo]{yun2019cutmix}
Sangdoo Yun, Dongyoon Han, Seong~Joon Oh, Sanghyuk Chun, Junsuk Choe, and Youngjoon Yoo.
\newblock Cutmix: Regularization strategy to train strong classifiers with localizable features.
\newblock In \emph{Proceedings of the IEEE/CVF international conference on computer vision}, pages 6023--6032, 2019.

\bibitem[Zagoruyko and Komodakis(2016)]{zagoruyko2016paying}
Sergey Zagoruyko and Nikos Komodakis.
\newblock Paying more attention to attention: Improving the performance of convolutional neural networks via attention transfer.
\newblock \emph{arXiv preprint arXiv:1612.03928}, 2016.

\bibitem[Zhang et~al.(2022)Zhang, Zhang, Liu, Weller, Sch{\"o}lkopf, and Xing]{zhang2022towards}
Hanlin Zhang, Yi-Fan Zhang, Weiyang Liu, Adrian Weller, Bernhard Sch{\"o}lkopf, and Eric~P Xing.
\newblock Towards principled disentanglement for domain generalization.
\newblock In \emph{Proceedings of the IEEE/CVF Conference on Computer Vision and Pattern Recognition}, pages 8024--8034, 2022.

\bibitem[Zhang et~al.(2017)Zhang, Cisse, Dauphin, and Lopez-Paz]{zhang2017mixup}
Hongyi Zhang, Moustapha Cisse, Yann~N Dauphin, and David Lopez-Paz.
\newblock mixup: Beyond empirical risk minimization.
\newblock \emph{arXiv preprint arXiv:1710.09412}, 2017.

\bibitem[Zhao et~al.(2022)Zhao, Cui, Song, Qiu, and Liang]{Zhao2022Decoupled}
Borui Zhao, Quan Cui, Renjie Song, Yiyu Qiu, and Jiajun Liang.
\newblock Decoupled knowledge distillation.
\newblock In \emph{Proceedings of the IEEE/CVF Conference on computer vision and pattern recognition}, pages 11953--11962, 2022.

\bibitem[Zhou et~al.(2021)Zhou, Yang, Qiao, and Xiang]{zhou2021domain}
Kaiyang Zhou, Yongxin Yang, Yu~Qiao, and Tao Xiang.
\newblock Domain generalization with mixstyle.
\newblock In \emph{International Conference on Learning Representations}, 2021.

\bibitem[Zhou et~al.(2022)Zhou, Zhang, Zang, Yang, Loy, and Liu]{zhou2022device}
Kaiyang Zhou, Yuanhan Zhang, Yuhang Zang, Jingkang Yang, Chen~Change Loy, and Ziwei Liu.
\newblock On-device domain generalization.
\newblock \emph{arXiv preprint arXiv:2209.07521}, 2022.

\bibitem[Zou et~al.(2023)Zou, Cao, Li, and Gu]{zou2023understanding}
Difan Zou, Yuan Cao, Yuanzhi Li, and Quanquan Gu.
\newblock Understanding the generalization of adam in learning neural networks with proper regularization.
\newblock In \emph{International Conference on Learning Representations}, 2023.

\end{thebibliography}

\appendix
\end{document}